\documentclass[10pt, conference, compsocconf]{IEEEtran}
\usepackage{cite}
\usepackage{amsmath,amssymb,amsfonts}
\usepackage[pdftex]{graphicx}
\usepackage{textcomp}
\usepackage{xcolor}
\def\BibTeX{{\rm B\kern-.05em{\sc i\kern-.025em b}\kern-.08em
    T\kern-.1667em\lower.7ex\hbox{E}\kern-.125emX}}

\usepackage{tikz,lipsum}
\usepackage[most]{tcolorbox}
\AtBeginDocument{%
  \providecommand\BibTeX{{%
    \normalfont B\kern-0.5em{\scshape i\kern-0.25em b}\kern-0.8em\TeX}}}
\usepackage{booktabs}
\usepackage{graphicx}
\usepackage[utf8]{inputenc}
\usepackage{hyperref}
\usepackage{balance}
\usepackage{url}
\usepackage{multirow}
\usepackage{caption}
\usepackage{wrapfig}
\usepackage{color,soul}
\usepackage{fancyvrb}
\usepackage{eepic}
\usepackage{enumitem}
\usepackage{caption}
\usepackage{subcaption}
\usepackage{graphicx}
\usepackage{soul}
\usepackage{booktabs}
\usepackage{booktabs, multirow} % for borders and merged ranges
\usepackage{array}
\newcolumntype{P}[1]{>{\centering\arraybackslash}p{#1}}
\def\shorten{\looseness=-1} 

\usepackage{algorithm,algpseudocode}
\algnewcommand{\To}{\textbf{To }}
\algnewcommand\Input{\item[\textbf{Input:}]}%
\algnewcommand\Output{\item[\textbf{Output:}]}%

\algnewcommand\algorithmicworker{\textbf{Worker}}
\algnewcommand\Worker{\algorithmicworker{} }
\newtheorem{definition}{Definition}[section]
\newcommand{\sysName}{KG-TOSA} 

\usepackage[hyphenbreaks]{breakurl}

\newcommand{\RDFTYPE}[3]{\ensuremath{{\langle}\texttt{#1},} \texttt{#2}, \ensuremath{\texttt{#3}{\rangle}}}

\newcommand{\myNum}[1]{(\emph{#1})}

\newcommand{\nsstitle}[1]{\noindent\textup{\textbf{#1}}}
\newcommand{\highlightedReply}[2]{{\color{black}{#1}}{}}
\begin{document}

\title{Task-Oriented GNNs Training on Large Knowledge Graphs for Accurate and Efficient Modeling}

\author{
\IEEEauthorblockN{Hussein Abdallah}
\IEEEauthorblockA{\textit{Concordia University, }\\
% Canada \\
hussein.abdallah@mail.concordia.ca}
\and
\IEEEauthorblockN{Waleed Afandi}
\IEEEauthorblockA{\textit{Concordia University, }\\
% Canada \\
waleed.afandi@concordia.ca}
\and
\IEEEauthorblockN{Panos Kalnis}
\IEEEauthorblockA{\textit{KAUST, }\\
% Canada \\
panos.kalnis@kaust.edu.sa}
\and
\IEEEauthorblockN{Essam Mansour}
\IEEEauthorblockA{\textit{Concordia University, }\\
% Canada \\
essam.mansour@concordia.ca}
}

\maketitle
\begin{abstract}
\sloppy
A Knowledge Graph (KG) is a heterogeneous graph encompassing a diverse range of node and edge types. Heterogeneous Graph Neural Networks (HGNNs) are popular for training machine learning tasks like node classification and link prediction on KGs. However, HGNN methods exhibit excessive complexity influenced by the KG's size, density, and the number of node and edge types. AI practitioners handcraft a subgraph of a KG $G$ relevant to a specific task. We refer to this subgraph as a task-oriented subgraph (TOSG), which contains a subset of task-related node and edge types in $G$. 
Training the task using TOSG instead of $G$ alleviates the excessive computation required for a large KG. Crafting the TOSG demands a deep understanding of the KG's structure and the task's objectives. Hence, it is challenging and time-consuming. 
This paper proposes {\sysName}, an approach to automate the TOSG extraction for task-oriented HGNN training on a large KG. 
In {\sysName}, we define a generic graph pattern that captures the KG's local and global structure relevant to a specific task. 
We explore different techniques to extract subgraphs matching our graph pattern: namely \myNum{i} two techniques sampling around targeted nodes using biased random walk or influence scores, and \myNum{ii} a SPARQL-based extraction method leveraging RDF engines' built-in indices. Hence, it achieves negligible preprocessing overhead compared to the sampling techniques.  
We develop a benchmark of real KGs of large sizes and various tasks for node classification and link prediction.
Our experiments show that {\sysName} helps state-of-the-art HGNN methods reduce training time and memory usage by up to 70\% while improving the model performance, e.g., accuracy and inference time.\shorten

% , and reducing the model size 
% for better accuracy, smaller model size, and less inference time.\shorten 
%and inference time
\end{abstract}

% We develop  a benchmark of real KGs of large sizes, up to hundred millions of triples and hundreds of node and edge types with various tasks for node classification and link prediction. 
\maketitle

\section{Introduction}
\label{sec:intro} 

A knowledge graph (KG) is a heterogeneous graph that includes nodes representing different entities ($\mathcal{V}$) of $|\mathcal{C}|$ classes, e.g., \textit{Paper} and \textit{Author} and edge types representing $|\mathcal{R}|$ relations, e.g., \textit{publishedIn} and \textit{write}, between these entities. $\mathcal{C}$ and $\mathcal{R}$ are the sets of node and edge types in the KG, respectively. Real-world KGs, such as Yago~\cite{KG_Yago4} and Wikidata \cite{KG_Wikidata}, have hundreds to thousands of node/edge types.
Heterogeneous graph neural networks (HGNNs) have emerged as a powerful tool for analyzing KGs by defining real-world problems as node classification, such as recommender systems~\cite{gnnRS}, and entity alignment~\cite{gnnCrossLingual}, or link prediction, such as drug discovery~\cite{kgnn} and fraud detection~\cite{fraudDet} tasks.

HGNN methods adopt mechanisms from GNNs designed for homogeneous graphs, such as multi-layer graph convolutional networks (GCNs) and sampling in mini-batch training~\cite{GraphSAINT, Shadow-GNN, FastGCN, MHGCN, MorsE,LHGNN}.  
Unlike GNNs on homogeneous graphs, the number of GCN layers in HGNNs is associated with the number of edge types or metapaths in the heterogeneous graph to model the distinct relationships effectively. A metapath captures distinct semantic relationships (edge types) between node types. For example, the metapath Author-(write)-Paper-(publishedIn)-Venue (APV) describes the semantic relationships of an author who published a paper in a particular venue. 
In mini-batch HGNN training, the incorporated sampling mechanisms aim to obtain a subgraph (a subset of nodes and edges) that captures the representative structure of the original KG in general. The sampling is performed without considering node and edge types, where some specific types may have negligible or no effect in training a particular task. For example, a DBPedia~\footnote{The DBPedia KG contains information about movies and academic work.} subgraph, including instances of APV does not contribute to training a model predicting a movie genre. Hence, HGNN methods are not optimized for identifying subgraphs of a smaller subset of types to reduce the number of GCN layers in training a specific task. 

The space complexity of training HGNNs includes the memory required to store the graph structures, associated features, and multi-layer GCNs. Additionally, the time complexity includes the sampling process, in the case of mini-batch training, and the message-passing iterations needed to aggregate the embeddings of neighbouring nodes in each GCN.
That leads to computationally expensive memory usage and training time with large KGs, where the complexity~\footnote{The actual complexity varies depending on the specific implementation.} is influenced by the KG's size, density,  $|\mathcal{R}|$, and  $|\mathcal{C}|$~\cite{SeHGNN, GraphSAINT, Shadow-GNN}.
To alleviates the excessive computation on a large KG, AI practitioners manually handcraft a subgraph of a subset of $\mathcal{R}$ and $\mathcal{C}$ from a given KG for training HGNNs~\cite{OGB,HGNN_Benchmark}. We refer to this subgraph as a task-oriented subgraph (TOSG). Training the task using TOSG instead of the original KG helps HGNN methods reduce training time and memory usage. For example, the open graph benchmark (OGB)~\cite{OGB-LSC, OGB} provided OGBN-MAG, a subgraph of 21M triples out of 8.4B triples $\approx 0.2\%$ that is extracted from the Microsoft Academic Graph (MAG)~\cite{MAG} and is relevant to the node classification task for predicting a paper venue ($PV$). OGBN-MAG contains nodes and edges of only four types. 
% handcrafted by an AI practitioner for this specific task.

\begin{figure}[t]
\vspace*{-1ex}
  \centering
  \includegraphics [width=\columnwidth]{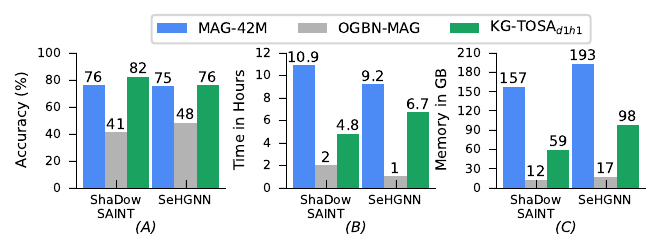}
  \vspace*{-4ex}
  \caption{(A) Accuracy (higher is better), (B) Training-Time (lower is better), (C) Training-Memory (lower is better). Training a node classification task ($PV$) to predict the paper venue using ShaDowSAINT~\cite{Shadow-GNN} and SeHGNN ~\cite{SeHGNN} on a MAG graph with 42M vertices ($MAG$$-$$42M$). The handcrafted task-oriented subgraph ($OGBN$$-$$MAG$) from $MAG$$-$$42M$ trades accuracy to reduce time and memory usage. Our {\sysName}$_{d1h1}$ task-oriented subgraph is extracted automatically from $MAG$$-$$42M$ for $PV$ to reduce time and memory consumption while improving accuracy.\shorten} 
 \label{fig:motivation}  
  \vspace*{-6ex}
\end{figure}

To demonstrate the effectiveness of this methodology, 
we trained the $PV$ task on a super-set of OGBN-MAG that contains 166M triples $\approx 2\%$ of MAG KG with 42M vertices, 58 node types and 62 edge types; we refer to it as ($MAG-42M$). The $OGBN$$-$$MAG$~\cite{OGB} is a subgraph of $MAG$$-$$42M$ relevant to $PV$, i.e., task-oriented subgraph for $PV$. 
We used the state-of-the-art GNN methods, namely ShaDowSAINT~\cite{Shadow-GNN} and SeHGNN ~\cite{SeHGNN}. As shown in Figure~\ref{fig:motivation}, $OGBN$$-$$MAG$ trades the accuracy to help both methods reduce time and memory usage in training the task. 
Furthermore, crafting the TOSG demands a deep understanding of the KG's structure and the task's objectives. Hence, the TOSG extraction is a challenging and time-consuming process and does not guarantee to improve the model performance.  

This paper proposes {\sysName}~\footnote{
{\sysName} stands for KG \underline{T}ask \underline{O}riented \underline{Sa}mpling. It is available at \url{https://github.com/CoDS-GCS/KGTOSA}}
% \url{https://gitfront.io/r/HGNN/LWYY16iUMSuE/KGTOSA/}}
, an approach to automate the TOSG extraction for task-oriented HGNN training on a large KG. In {\sysName}, we define a generic graph pattern that effectively captures both the local and global structure of a KG for a specific task. Our definition is grounded in the theoretical foundation of training HGNNs and considering crucial factors,  such as data sufficiency and graph topology.
The objective of our graph pattern is to maximize the diversity of neighbour node types while minimizing the average depth between non-target and target vertices.
% \textcolor{blue}{Our graph pattern balances the ratio of nodes targeted by the task} and the diversity of node and edge types within the subgraph.
We preserve the local context by identifying an initial set of nodes targeted by the task. Then, we expand our selection to include neighbouring nodes within a specified distance of $h$ hops. Our graph pattern interlinks the local context around each target node to generate a larger subgraph of node and edge types globally related to the task. This approach allows us to effectively capture the intricate relationships and dependencies between different entities in the KG to enhance the representation of the task-specific knowledge captured in the TOSG.\shorten

We explore different techniques to extract subgraphs matching our graph pattern. Two techniques sample around targeted nodes using biased random walk or influence scores. We also propose a SPARQL-based extraction method leveraging RDF engines' built-in indices. Our biased random walk sampling performs random walks starting from nodes that match the target vertices in the task. In our influence-based sampling approach, we utilize a Personalized PageRank (PPR) score to measure the relevance or importance of nodes in the KG to the task. However, it is essential to note that these sampling methods can incur a significant extraction overhead, particularly with a large KG. 
This excessive overhead may overshadow the potential time savings in training using the TOSG generated by both sampling methods instead of the original KG. To mitigate this overhead, we propose a SPARQL-based method that leverages built-in indices in RDF engines. Hence, our SPARQL-based method automates the TOSG extraction with negligible preprocessing overhead. 
Figure~\ref{fig:motivation} motivates the need for our {\sysName} approach, which helps both ShaDowSAINT and SeHGNN \myNum{i} reduce the overall maintenance cost of the multi-layer GCNs, and \myNum{ii} improve the modelling performance, including accuracy and inference time.\shorten

% by showing the performance improvement in both ShaDowSAINT and SeHGNN using the TOSG extracted by the SPARQL-based method.
% By adopting our SPARQL-based method, HGNN methods can achieve two benefits: \myNum{i} reducing the overall maintenance cost of the multi-layer GCNs, and \myNum{ii} improving the modeling performance, including accuracy and inference time. 

For evaluating GNN methods on node classification and link prediction tasks, existing benchmarks, such as OGB \cite{OGB}, OGB-LSC \cite{OGB-LSC}, and Open-HGNN \cite{HGNN_Survey}, include KGs of few node/edge types. Unlike these benchmarks, we develop a benchmark of real KGs of large sizes, up to hundred millions of triples and hundreds of node and edge types with various tasks for node classification and link prediction. Hence, our benchmark presents challenging settings for the state-of-the-art (SOTA) HGNN methods. 
Our benchmark encompasses KGs from the academic domain, such as MAG~\cite{MAG} and DBLP~\cite{KG_DBLP}, as well as two versions of YAGO as a general-purpose KG of different sizes, namely YAGO-4~\cite{KG_Yago4} and YAGO-3~\cite{Yago3-10}. 
% We also have formulated various graph machine learning tasks, which have been translated into node classification and link prediction problems. 
%
We integrated {\sysName} with SOTA GNN methods, namely SeHGNN~\cite{SeHGNN}, GraphSAINT~\cite{GraphSAINT}, ShaDowSAINT~\cite{Shadow-GNN}, RGCN~\cite{RGCN},  MorsE~\cite{MorsE} , and LHGNN \cite{LHGNN}.  Our comprehensive evaluation using large real KGs and various ML tasks shows that {\sysName} enables GNN methods to converge faster and use less memory for training. Moreover, {\sysName} empowers GNN methods to enhance and attain comparable levels of accuracy.\shorten  

In summary, our contributions are:
 \vspace*{-2ex}
\begin{itemize}

\item the first approach ({\sysName}) for task-oriented HGNN training based on our generic graph pattern identifying a KG's local and global context related to a specific task, Section~\ref{sec_approach}. \shorten 

\item three mechanisms for automating the TOSG extraction using our generic graph pattern based on task-oriented sampling and SPARQL-based query extraction. Our SPARQL-based method outperforms the task-oriented sampling in time and memory usage, Section~\ref{sec_Algo}.     

\item a comprehensive evaluation using our KG benchmark and six SOTA GNN methods. Our KG benchmark includes four real KGs, six node classification tasks, and three link-prediction tasks. The KGs have up to 42 million nodes and 400 million edges. {\sysName} achieves up to 70\% memory and time reduction while improving the model performance for better accuracy, smaller model size, and less inference time, Section~\ref{sec:expermients},
% e.g., accuracy and inference time.\shorten %
\end{itemize}

% The remainder of this paper starts with Section~\ref{sec:background} to clarify the necessary background. Section~\ref{sec_approach} presents the {\sysName} approach based on our generic graph pattern. Sections~\ref{sec:ben} and~\ref{sec:expermients} introduce our benchmark and evaluation. We highlight related work in Section~\ref{sec:relatedwork} and conclude the paper in Section~\ref{sec:conclusion}.
\section{Background}
\label{sec:background} 

A Knowledge Graph ($KG$) is a graph representation of heterogeneous information defined as: 
\begin{definition}[Knowledge Graph]
\label{def:kg} 
Let $\mathcal{V}$ be a set of vertices representing entities, $\mathcal{C}$ be a set of classes (node types) that correspond to the entities in $\mathcal{V}$, $\mathcal{R}$ be a set of relations, and $\mathcal{L}$ be a set of literal values. The knowledge graph ($KG$) is a directed multigraph $KG = (\mathcal{V}, \mathcal{C}, \mathcal{L}, \mathcal{R}, \mathcal{T})$, where $\mathcal{V} = \left\{v: type(v) \in \mathcal{C}\right\}$ and $\mathcal{T} = \left\{(s, p, o): s \in \mathcal{V}, p \in \mathcal{R}, o \in \mathcal{V} \lor o \in \mathcal{L}\right\}$. Each $(s, p, o)$ in $\mathcal{T}$ represents an edge between a subject $s$ and an object $o$ via a predicate $p$.
\end{definition}

% \begin{definition}[Knowledge Graph]
% \label{def:kg} 
% Let $\mathcal{V}$ be a set of vertices representing entities, such as James Nicholas Gray, $\mathcal{C}$ be a set of classes (node types) that correspond to the entities in $\mathcal{V}$, such as \textit{person} (James is a person), $\mathcal{R}$ be a set of relations, such as \textit{author of}, and $\mathcal{L}$ be a set of literal values, such as strings, numbers, and dates. The knowledge graph ($KG$) is a directed multigraph $KG = (\mathcal{V}, \mathcal{C}, \mathcal{L}, \mathcal{R}, \mathcal{T})$, where $\mathcal{V} = \left\{v: type(v) \in \mathcal{C}\right\}$ and $\mathcal{T} = \left\{(s, p, o): s \in \mathcal{V}, p \in \mathcal{R}, o \in \mathcal{V} \lor o \in \mathcal{L}\right\}$. Each $(s, p, o)$ in $\mathcal{T}$ represents an edge between a subject $s$ and an object $o$ via a predicate $p$.\shorten
% \end{definition}

\begin{figure*}[t]
\vspace*{-3ex}
     \centering
     \begin{subfigure}[b]{0.34\textwidth}
         \centering        \includegraphics[width=\textwidth]{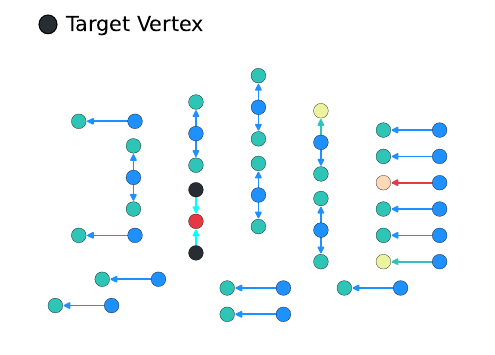}
         \vspace*{-5ex}
         \caption{YAGO-30M (CreativeWork-Genere) 15.25\%}
         \label{fig:YAGO-30M-RW}
     \end{subfigure}
     \hfill
     \begin{subfigure}[b]{0.32\textwidth}
         \centering         \includegraphics[width=\textwidth]{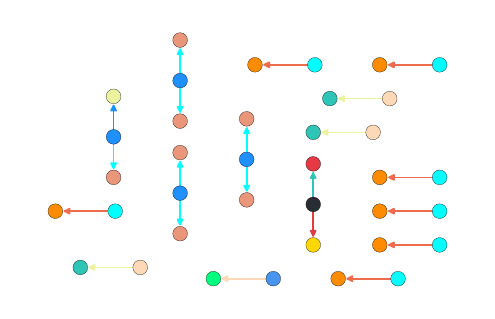}
         \vspace*{-5ex}
         \caption{MAG-42M (Paper-Venue) 73.79\%}
         \label{fig:MAG-42M-RW}
     \end{subfigure}
     \hfill
     \begin{subfigure}[b]{0.32\textwidth}
         \centering         \includegraphics[width=\textwidth]{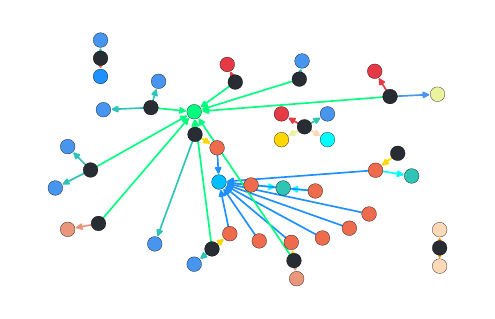}
         \vspace*{-5ex}
         \caption{DBLP-15M (Paper-Venue) 81.79\%}
         \label{fig:DBLP-15M-RW}
     \end{subfigure}
     % \vspace*{-1ex}
        \caption{Examples of subgraphs generated by the uniform random walk (URW) sampler in GraphSAINT for different GNN tasks. The black vertices indicate the target vertices. The same colour means the same vertex or edge type. This sampling method does not guarantee to \myNum{i} include enough representation of target vertices ($\mathcal{V}_T$) and \myNum{ii} exclude vertices disconnected from $\mathcal{V}_T$ as these vertices do not contribute to the embeddings of $\mathcal{V}_T$.\shorten}
         \vspace*{-1ex}
\label{fig:RW_subgraphQuality}
\vspace*{-2ex}
\end{figure*}

\subsection{HGNNs on KGs}
 \label{bg_GNN}

Existing HGNN methods build upon GNN techniques designed for homogeneous graphs by incorporating multi-layer GCNs and employing sampling in mini-batch training~\cite{GraphSAINT, Shadow-GNN, FastGCN, MHGCN, MorsE}. In HGNNs, the number of GCN layers is associated with the number of metapaths in the heterogeneous graph. 
Metapaths can range from a single edge type and one-hop to paths with multiple hops and edge types~\cite{SeHGNN,LHGNN}. For a
$KG = (\mathcal{V}, \mathcal{C}, \mathcal{L}, \mathcal{R}, \mathcal{T})$, a metapath is defined as a sequence of relations across node types in  $\mathcal{C}$ in the form of $c_1 \xrightarrow[\text{}]{\text{$r_1$}}c_2 \xrightarrow[\text{}]{\text{$r_2$}}...\xrightarrow[\text{}]{\text{$r_h$}}c_{h+1}$. This leads to composite relations (metapaths), which construct all possible combinations of $\mathcal{C}$ and $\mathcal{R}$ for hop $h$ based on the KG schemas.

Relational Graph Convolutional Network (RGCN)\cite{RGCN} is based on a metapath of one hop and edge type. RGCN uses a stack of GCNs, where each GCN is dedicated to a metapath involving one specific edge type. Various GNN methods, such as GraphSAINT\cite{GraphSAINT}, Shadow-GNN~\cite{Shadow-GNN}, MorsE~\cite{MorsE}, and FastGCN~\cite{FastGCN}, have adapted RGCN to support heterogeneous graphs. We refer to this category of methods as RGCN-based HGNN methods. Another category of HGNN methods supports metapaths involving multiple hops of different edge types, e.g., SeHGNN~\cite{SeHGNN}, MHGCN~\cite{MHGCN}, and FastGTN~\cite{FastGTN}.
% , and NARS~\cite{NARS}. 
We refer to this category as metapath-based methods~\cite{PathSim}. 

In both categories, the initial node embeddings are iteratively aggregated with embeddings received from neighboring nodes connected to a specific relation until the embeddings of all nodes converge. For instance, in RGCN-based methods, the final embedding of a node is obtained through two aggregations:  an outer aggregation over each relation type and an inner aggregation over neighbouring nodes of a specific relation and defined by RGCN~\cite{RGCN} as follows:

\begin{equation}
\label{eq_rgcn}
h_i^{(l+1)}= \sigma(\sum_{r \in R}\sum_{j \in N_i^r}\frac{1}{c_{i,r}}W_r^{(l)}h_j^{(l)}+W_0^{(l)}h_j^{(l)}) 
\end{equation}

where $l$ is an RGCN layer, $h^{(l+1)}_j$ is the hidden embedding of node $j$ at layer $l+1$, $\sigma$ is element-wise activation function, $N_i^r$ denotes the set of neighbour indices of node $i$ under relation $r \in \mathcal{R}$, $c_{i,r}$ is a problem-specific normalization constant that can either be learned or chosen in advance (such as $c_{i,r}=|N_i^r|$), $W^{(l)}_r$ is the weight matrix for relation $r$ at layer $l$, and $W^l_0$ is the initial weight matrix at layer $l$.\shorten

% \textcolor{orange}{Unlike HGNN methods, knowledge graph embeddings (KGE) methods, such as TransE~\cite{Realistic_Eval_KGEs} and others in \cite{PyKEEN} learn scoring functions to distinguish valid triples from invalid ones. KGE models do not account for multi-hop neighbour contexts and do not support inductive inference\cite{Inductive_LP,Inductive_QE}. Hence, KGE methods are not considered a subclass of GNNs and fall outside the scope of this paper.\shorten}

% \textcolor{orange}{Unlike HGNN methods, knowledge graph embeddings (KGE) methods ~\cite{Realistic_Eval_KGEs} do not account for multi-hop neighbour aggregations and do not support inductive inference\cite{Inductive_LP,Inductive_QE}. Hence, KGE methods fall outside the scope of this paper.\shorten}

\subsection{Sampling Techniques Adapted in HGNNs}
% HGNN mini-batch training methods use different sampling techniques to split each training epoch into multiple mini-batches on a subset of the KG~\cite{SamplingSurvey}.\shorten 

HGNN mini-batch sampling methods perform sampling method on a KG then train on a set of sampled subgraphs ~\cite{SamplingSurvey}.\shorten 

% \nsstitle{Sampling in RGCN-based Methods:}
% Several RGCN-based GNN Methods, such as GraphSAINT~\cite{GraphSAINT}, MorsE~\cite{MorsE} and ShaDowSAINT~\cite{Shadow-GNN},

\nsstitle{Sampling in RGCN-based Methods:}
Several RGCN-based GNN Methods in ~\cite{GraphSAINT, MorsE,Shadow-GNN},
adopt different sampling techniques. These techniques are originally designed for homogeneous graphs to reduce only the number of nodes. Hence, they do not guarantee to reduce node and edge types, which is required in extracting the TOSG. These sampling techniques can be classified into three main groups: node-wise, layer-wise, and recently subgraph sampling. As Node-wise and Layer-wise sampling techniques suffer from scalability to large KGs \cite{SamplingSurvey}. Recent GNN methods, such as GraphSAINT\cite{GraphSAINT}, Shadow-GNN~\cite{Shadow-GNN}, utilize subgraph-based sampling techniques to sample a subgraph from the original graph $G$ in each epoch. The sampling method ensures that the sampled subgraph ${G^\prime}$ preserves the general structure of $G$.  GraphSAINT subgraph sampler uses a uniform random-walk sampler (URW) by default to randomly select a set of initial root nodes and performs a random walk of length $h$ from each root node to its neighbours. URW collects all the nodes and edges that are reached in the whole process that forms a sample graph $G^\prime(\mathcal{V}^\prime,\mathcal{R}^\prime)$. 
Through URW the high degree nodes have higher probability to be visited without consideration for node/edge types \cite{ibmb}. GraphSAINT further applies normalization techniques during the training to prevent the bias in the induced sub-graphs.\shorten

\nsstitle{Sampling for Metapaths:}
The idea is to keep balanced nodes count per node type $c$ 
while generating the metapaths instances w.r.t the high variability of KG nodes/edges types. Importance-sampling strategy as in \cite{HGT} is used to reduce the variance in metapaths instances and reduce the number of sampled nodes/edges.
Constructing subgraphs for all possible metapaths is a time-intensive process \cite{NARS}. Additionally, aggregating these metapaths into a metapath graph also incurs a significant amount of memory overhead \cite{MHGCN}. SeHGNN~\cite{SeHGNN} optimizes the metapath sampling and features aggregation cost by performing the GNN neighbour aggregation only once in the pre-processing stage. \shorten

\subsection{GNN Tasks on KGs}
GNNs are particularly well adapted in real-world applications for node classification (NC) and link prediction (LP) tasks~\cite{GNN_methods_and_applications_2021}.  Examples of NC tasks are predicting the community to which a node belongs in a social network~\cite{GNN_NC_Survey} or a venue for publishing a paper~\cite{OGB}.\shorten 

\begin{definition}[Node Classification]
\label{def:NC} 
For a given $KG = (\mathcal{V}, \mathcal{C}, \mathcal{L}, \mathcal{R}, \mathcal{T})$, a node classification task $NC(KG, \mathcal{V}_T, c_T)$ aims to predict labels for each target vertex $v_t \in \mathcal{V}_T$, where $v_t$ has the node type $c_T$. The predicted labels can be either a single-label or a multi-label, depending on the problem setting.\shorten
\end{definition}

A single-label NC task predicts only one label from a set of mutually exclusive labels. The task objective is to predict the most appropriate label for each node, e.g., a venue of a paper is predicted from a set of venues. In contrast, a multi-label NC task predicts multiple labels simultaneously. The task aims to predict the presence or absence of multiple labels for each node~\cite{GNN_NC_Survey}, e.g., predicting keywords of a paper.\shorten

LP tasks predict the potential or missing link between existing nodes in a graph, e.g., affiliations of an author. LP is formalized either as predicting the missing entities (vertices) or predicting the missing relations~\cite{LP_KG_Survey}. In this paper, we consider the missing entities task where we predict the correct vertex that completes ⟨$v_t$,p,?⟩ or ⟨?,p,$v_t$⟩. We refer to the known vertex as the target vertex and perform link prediction as follows:\shorten

\begin{definition}[Link Prediction]
\label{def:LP} 
For a given $KG = (\mathcal{V}, \mathcal{C}, \mathcal{L}, \mathcal{R}, \mathcal{T})$ and for every predicate (edge type) $p_T \in \mathcal{R}$, a link prediction task $LP(KG, \mathcal{V}_T, C_T, p_T)$ aims to predict a set of potential vertices that could be connected to the vertex $v_t \in \mathcal{V}_T$ via the predicate $p_T$, where $v_t$ belongs to the vertex set $\mathcal{V}_T$ and has the type $c_T \in C_T$.
\end{definition}

Existing benchmarks for HGNN methods, such as \cite{OGB, OGB-LSC, OpenHGNN, HGNN_Benchmark}, use heterogeneous graphs of a few node/edge types. For example, the subgraph MAG240M in OGB~\cite{OGB-LSC} includes three node and edge types. These benchmarks defined different NC and LP tasks to be modelled on these graphs.  Real KGs contain hundreds to thousands of node and edge types such as Wikidata \cite{KG_Wikidata} containing 1,301 and 10,012 node and edge types and Yago~\cite{KG_Yago4} containing 8,902 and 156 node and edge types, respectively. There is a need to benchmark using heterogeneous graphs with up to hundreds of types.\shorten 

\section{The {\sysName} Extraction Approach}
\label{sec_approach}

\highlightedReply{The core contribution of our approach is based on our comprehensive study of data sufficiency and graph topology on HGNNs training. Our findings enable us to discover and design a generic graph pattern for extracting a TOSG from a KG for a specific task. This section highlights these findings and the graph pattern.\shorten}{R1.O2}

% Our comprehensive study of data sufficiency and graph topology on HGNNs training enables us to discover a generic graph pattern for extracting a TOSG from a KG for a specific task. This graph pattern is the core contribution of our approach.\shorten    

\subsection{Characteristics of Effective HGNNs Training}
\label{sec:chars}

Our study explored the factors contributing to good performance and strong generalization in training HGNNs for a specific task. We focused on understanding the characteristics of the graph structure that enable HGNNs to generate distinguished embeddings for vertices targeted by a graph-related task. The learned embeddings are then utilized in a downstream task, such as node classification or link prediction. Our study included two main characteristics, namely data sufficiency and graph topology.
In our study,  we used GraphSAINT and its Uniform Random Walk (URW)  sampler to investigate the quality of the sampled subgraph using three different tasks in three different KGs, as shown in Figure \ref{fig:RW_subgraphQuality}.(a,b,c). URW samples these subgraphs with a walk-length (no. of hops) of $h=2$ and an initial set of nodes of 20 for training a GNN task.\shorten

\nsstitle{Data Sufficiency:} 
Adequate and representative data is essential for training HGNNs effectively~\cite{EffectofTrainingData, HetGNN}. Sufficient data ensures that the model has access to diverse graph structures and informative node/edge attributes relevant to the task at hand. This allows the training process to capture meaningful patterns and relationships. 
% \textcolor{orange}{With appropriate data, HGNNs can learn to generate high-quality embeddings that encode important features and enable accurate predictions in downstream tasks.} 
Existing HGNN methods consider the whole KG in training or sample uniformly a representative graph structure.
Suppose the training data includes an insufficient number or low ratio of target vertices. This leads to training batches (subgraphs) with a few target vertices (black vertices shown in Figure~\ref{fig:RW_subgraphQuality}). The subgraphs in Figure~\ref{fig:RW_subgraphQuality}.(a) and Figure~\ref{fig:RW_subgraphQuality}.(b) are examples of this case. This limited representation of target vertices adversely impacts the model performance with a slower convergence rate, i.e., increases the training time. Figure~\ref{fig:RW_subgraphQuality} shows that these methods do not guarantee a diverse graph structure around the target vertices. For example, the target vertices are connected to a limited number of vertices of other types.\shorten
% , as shown Figure~\ref{fig:RW_subgraphQuality}.

\nsstitle{Graph Topology:} We also investigate the graph topology. The arrangement and connectivity patterns of nodes and edges profoundly impact the quality of  embeddings and the efficiency of generating them in HGNNs. The graph topology determines how information propagates and diffuses across the graph during message passing. It influences the reachability of nodes, the diversity of neighbour node types, and the local/global context available for each vertex. By considering the graph topology, HGNNs can effectively capture and leverage the structural dependencies to generate embeddings that encapsulate the relevant information for the given task. 

Suppose the training data includes vertices disconnected from the target vertices. Then, a GNN method will consume unnecessary aggregation iterations to calculate embeddings that will not affect the final embeddings of the target vertices. For example, subgraphs in Figure~\ref{fig:RW_subgraphQuality}.(a) and (b) include several vertices that are disconnected from target vertices. 
Moreover, the training data may include vertices that are far from the target vertices. For example, the subgraph in Figure~\ref{fig:RW_subgraphQuality}.(c) includes distanced vertices, e.g., up to 4 hops, from a target vertex. This leads to the over-smoothing problem~\cite{DropEdge, GNN_Over_Smoothing}, i.e., generating indistinguishable embeddings. 
Irrelevant or less related vertices have a diminished impact on model generalization and potentially hinder performance. Furthermore, including node/edge types unrelated to the task's target nodes can introduce unnecessary complexity to the GNN model layers. This increased complexity contributes to more model parameters and prolongs the training and inference time. 

\begin{figure}[t]
\vspace*{-3ex}
 \centering  \includegraphics[width=\columnwidth]{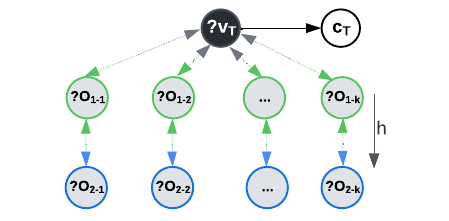}
  \vspace*{-4ex}
  \caption{The TOSG's generic graph pattern is based on two parameters: \myNum{i} the direction (outgoing and incoming) predicates, and \myNum{i} the number of hops.
  }
  \label{fig:node-BGP-samplers}
  \vspace*{-3ex}
 \end{figure} 

Diverse node neighborhoods help improve the efficacy of GNNs~\cite{Diversity_Matters}.  To assess graph structural diversity, we employ Shannon Entropy~\cite{Shannon_Entropy} to measure the variability in the number of neighbor node types per node, as follows:
\begin{equation}
\label{eq_shannon}
H(\mathcal{N}_t)= -\sum_{i=1}^{n} \mathbb{P}(\mathcal{N}_t(i)) \cdot \log_2(\mathbb{P}(\mathcal{N}_t(i))) 
\end{equation}
where $\mathbb{P}(\mathcal{N}_t(i))$ represents the probability of the count of neighbour node types for a node $i$ in a subgraph with $n$ nodes. A higher $H(\mathcal{N}_t)$ indicates increased diversity in the type of neighbouring nodes. This diversity enhances the graph's structure to enable more effective learning of node representations.\shorten

Even methods that utilize full-batch approaches, such as RGCN, train on the whole KG and treat all vertices equally in the training process. These methods do not exploit the potential benefits of data sufficiency and the inherent structure of the graph topology. Consequently, these methods allocate resources toward vertices that have no or little impact on the desired task. That leads to wasted computational time and memory usage. Hence, training using the TOSG is crucial for optimizing model complexity, and improving training and inference efficiency for both mini- and full- batch training methods.\shorten

\begin{figure}[t]
\vspace*{-3ex}
 \centering  \includegraphics[width=0.85\columnwidth]{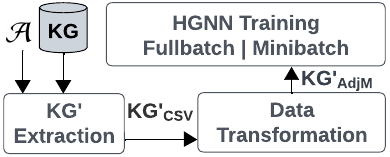}
  \vspace*{0ex}
  \caption{\highlightedReply{The {\sysName}  generic workflow. The TOSG ($KG'$) of Task $\mathcal{A}$ is extracted and transformed into an adjacency matrix. Then, the HGNN training performs either mini-batch or full-batch training. The size of $KG'$ is proportional to $\mathcal{V}_T$, the average degree of $\mathcal{V}_T$, and the average distance to $\mathcal{V}_T$. The smaller $KG'$, the faster the training pipeline.
  }{R2.O2, R3.O2}}
  \label{fig:KGTOSA_Pipeline}
  \vspace*{-3ex}
 \end{figure} 

\subsection{The {\sysName}'s Graph Pattern}
We introduce a graph pattern for a TOSG that is carefully designed based on our study of data sufficiency and graph topology. We define the TOSG as follows:

\begin{definition}[Task-oriented Subgraph for Training HGNNs]
\label{def:sgsampling} 
Given a knowledge graph $KG = (\mathcal{V}, \mathcal{C}, \mathcal{L}, \mathcal{R}, \mathcal{T})$, and a GNN task $\mathcal{A}$ targeting the set of vertices $\mathcal{V}_T$, there is a task-oriented subgraph of $KG$, denoted as $KG' = (\mathcal{V}', \mathcal{C}', \mathcal{L}', \mathcal{R}', \mathcal{T}')$. $KG'$ is a compact subset preserving the local and global graph structure of $KG$ relevant to $\mathcal{A}$. 
$KG'$ aims to maximize the neighbour node type entropy, where every non-target vertex is reachable to a vertex in $\mathcal{V}_T$, and minimize the average depth between non-target to target vertices.\shorten
\end{definition}

\begin{figure*}[t]
\vspace*{-3ex}
     \centering
       \begin{subfigure}[b]{0.33\textwidth}
         \centering         \includegraphics[width=\textwidth]{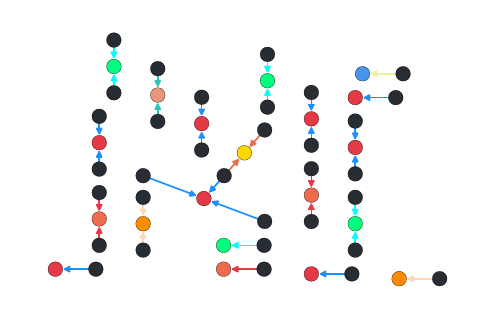}
         \vspace*{-4ex}
         \caption{YAGO-30M (CreativeW-Genere) 36.73\%}
         \label{fig:YAGO-30M-BRW}
     \end{subfigure}
     \hfill
     \begin{subfigure}[b]{0.31\textwidth}
         \centering         \includegraphics[width=\textwidth]{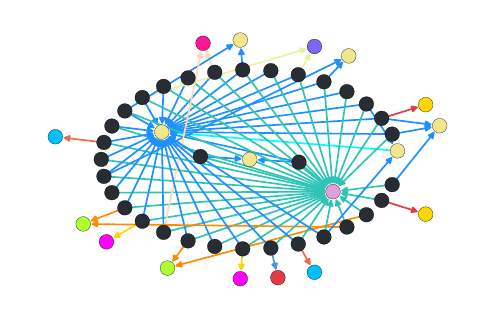}
         \vspace*{-4ex}
         \caption{MAG-42M (Paper-Venue) 75.33\%}
         \label{fig:MAG-42M-BRW}
     \end{subfigure}
     \hfill
     \begin{subfigure}[b]{0.31\textwidth}
         \centering         \includegraphics[width=\textwidth]{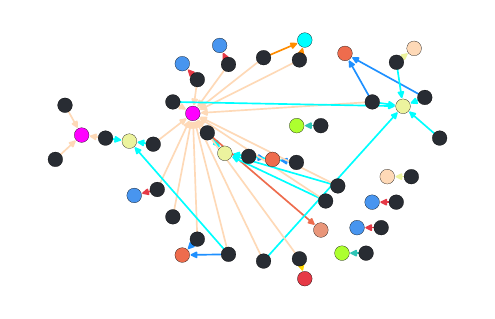}
         \vspace*{-4ex}
         \caption{DBLP-15M (Paper-Venue) 80.53\%}
         \label{fig:DBLP-15M-BRW}
     \end{subfigure}
     \vspace*{0ex}
     \caption{The subgraphs (a, b, and c) are generated by our biased random walk sampler. The black vertices indicate the target vertices. Node and edge types are colour-coded. Our approach leads to a higher ratio of target vertices w.r.t. RW in Figure~\ref{fig:RW_subgraphQuality} while the non-target vertices of different node/edge types are reachable to at least one vertex in $\mathcal{V}_T$.\shorten}        
\label{fig:subgraphQuality}
\vspace*{-3ex}
\end{figure*}

% Our graph pattern captures the general pattern of $KG'$ where its structure is centralized around the vertices targeted ($\mathcal{V}_T$) by the task at hand. For example, vertices of type Publications in the $PV$ task in Figure~\ref{fig:motivation}. 

In our graph pattern, we expand each vertex in $\mathcal{V}_T$ to its neighbouring vertices connected via incoming or outgoing predicates in a range of $h$ hops. Figure~\ref{fig:node-BGP-samplers} illustrates our graph pattern. This graph pattern helps to guarantee a high quality of the final embeddings generated for $\mathcal{V}_T$. These embeddings are obtained by aggregating the initial embeddings of $\mathcal{V}_T$ with those of its neighbouring vertices up to a specified hop distance $h$, where each iteration of aggregation reflects embeddings of neighbouring vertices at hop $h+1$ (e.g., neighbours of neighbours).\shorten 

Our graph pattern preserves the local structure for target nodes by considering the reachable incoming or outgoing neighbours to a certain depth $h$. Merging the $\mathcal{V}_T$ neighbours to construct the $KG^{\prime}$ will preserve the global structure for $\mathcal{V}_T$ by including the links between their neighbours which results in a deeper graph with node/edge distribution close to the original KG. \highlightedReply{Figure \ref{fig:KGTOSA_Pipeline} shows an end-to-end workflow of TOSG ($KG'$) extraction, transformation, and HGNN training.}{R2.O2}  Figure \ref{fig:subgraphQuality} illustrates examples of the subgraphs extracted using our graph pattern for the same tasks and KGs used in Figure~\ref{fig:RW_subgraphQuality}. In these subgraphs, all non-target vertices are reachable to at least one target vertex, and the target vertices are surrounded by more diverse sets of vertices and edges of different types.\shorten

\section{Our Graph Pattern in Practice: Algorithms}
\label{sec_Algo}
This section introduces three techniques for extracting subgraphs matching our graph pattern: biased random walk (BRW), influenced-based sampling (IBS), and a SPAQRL-based method. Each technique achieves this goal through different mechanisms (random walks, influence-based sampling, and SPARQL Basic Graph Pattern (BGP) matching), contributing to the overall objective outlined in Definition 3.1.
\highlightedReply{We adapted our graph pattern to extend the recent approaches applied to homogeneous graphs to develop our two main baselines: BRW and IBS. These baselines play a crucial role in evaluating the effectiveness of our SPARQL-based method.}{R2.O1}

\subsection{Biased Random Walk Sampling}
\label{sec:brw}

We developed the BRW technique, whose pseudocode is illustrated in Algorithm~\ref{alg:BRW_MS}. Our technique biases the random walk expansion
toward graph regions centered around the target vertices within a given walk length. 
We adopt the widely studied URW~\cite{GraphSAINT} to choose an initial set of vertices randomly from the set of $\mathcal{V}_T$.  This adoption maintains task-relevant nodes by excluding the task targets' disconnected nodes, as shown in Figure \ref{fig:subgraphQuality}.(a,b, and c). It also increases the number of target nodes to allow data sufficiency. The randomness in the walk, when biased towards the target vertices, contributes to maximizing neighbor node type entropy and ensuring reachability, i.e., better graph topology.\shorten

% \textcolor{blue}{URW is biased into high-degree nodes (Hubs) that are likely of a single node type such as a user node in the social network graph. Hub nodes are likely to have the same distribution of neighbor node types which decreases the sampled subgraphs neighbour-node-type’s entropy. BRW regularize the URW subgraph with starting nodes $\mathcal{V}_T$ of a single type. The different distribution of $\mathcal{V}_T$ neighbour node types increases the sampled subgraph neighbour-node-type’s entropy. 
% }
% \textcolor{blue}{Furthermore, It allows the sampling of diverse target-connected node types that improve the subgraph neighbour node types entropy}. \shorten

In Algorithm~\ref{alg:BRW_MS}, the function getInitialVertices randomly selects $bs$ vertices from $\mathcal{V}_T$, as shown in line 2. Then, a random walk sampler expands these initial vertices $\mathcal{V}_{initial}$ in $h$ hops to finally select a set of nodes $\mathcal{V}_s$ as shown in lines 3-6. The $KG^{\prime}$ subgraph is constructed in line 7 by including all edges between $\mathcal{V}_s$ nodes in the original KG. The BRW sampler preserves the local graph structure relevant to the task by considering connected vertices to $\mathcal{V}_T$ in performing the random walk. The subgraph extraction in line 7 interconnects all edges between these nodes to construct $KG^{\prime}$. This helps BWR interlink between different vertices located across the KG and related to the task.

Two factors dominate the complexity of BRW. First, the random walk complexity  $\mathcal{O}(h|\mathcal{V}_T|)$ in lines 4-6, where $h$ is the walk length that controls how far neighbour nodes to include in the sampled subgraph. Second, the subgraph extraction in line 7 with complexity  $\mathcal{O}(d|\mathcal{V}_s|)$, where d is the average degree of nodes in $\mathcal{V}_s$ that depends on the structure of the graph. 
\highlightedReply{As $d$ decreases, $\mathcal{V}_s$ generally increases in size, leading to a higher extraction cost for $KG^{\prime}$.}{R1.O1}\shorten
%When $d$ decreases, $\mathcal{V}_s$ tends to increase, which increases the extraction cost of $KG^{\prime}$. 
% Larger $|\mathcal{V}_s|$ and higher graph density increase the extraction cost of $KG^{\prime}$. 
%
Larger $\mathcal{V}_T$ and $h$ as well contributes to larger $|\mathcal{V}_s|$. Random walks are performed per $v_t$. So it could be easily parallelized. However, the dominating factor of Algorithm~\ref{alg:BRW_MS} is the subgraph extraction at line 7.  
% BRW samples the neighbouring nodes randomly. Hence, it may exclude neighbouring nodes representing important features to the target vertices. That can negatively affect the model's accuracy. However, t
The main advantage of BRW is the lightweight sampling complexity.\shorten 
\begin{algorithm}[t]
\caption{Biased Random Walk Sampling}
	\label{alg:BRW_MS}
        % \algsetup{linenosize=\small}
        \small
	\begin{algorithmic}[1]
             \Input \textit{Knowledge graph $KG$; GML task $\mathcal{A}$; Walk length $h$; batch size $bs$.}
    	\Output Knowledge graph \(KG^\prime\).
    	\Function{BRW\_MS}{$KG$, $\mathcal{A}$, $h$, $bs$}
        	\State $\mathcal{V}_{initial}  \leftarrow      
                \textrm{getInitialVertices(}bs, \mathcal{A}.\mathcal{V}_T)$     
        	\State $\mathcal{V}_s \leftarrow \mathcal{V}_{initial}$ 
        	\For {$v \in \mathcal{V}_{initial} $}
        		% \State $u \leftarrow v$
                   \State $\mathcal{V}_s \leftarrow \mathcal{V}_s \cup \text{randomWalkSampler }(KG,v,h)$
          	\EndFor
        	\State $ KG' \leftarrow \text{extractSubgraph}(\mathcal{V}_s, KG)$
    	\EndFunction
	\end{algorithmic} 
\end{algorithm} 
\vspace*{-1ex}

\subsection{Influenced-based Sampling}
\label{sec:ibs}

We also developed an influenced-based sampling  (IBS) technique that expands from target vertices to neighbouring nodes based on an influenced-based score. This score is to reflect the importance of a node to a particular target vertex. Our IBS technique adapts the PPR~\cite{PPR_Aprox} to approximate an influence score between the two nodes, where the influence score determines the local scope sensitivity of node $v$ on node $u$ as follows:\shorten

% A similar influenced-based technique, called IBMB~\cite{ibmb}, has been introduced for homogeneous graphs to sample representative subgraphs that could be used to train different tasks. IBMB used Personalized Pagerank (PPR) to calculate the influenced scores. IBMB does not support KGs and task-oriented HGNNs training.  
% Our IBS technique adapts the PPR~\cite{PPR_Aprox} to approximate an influence score between two the nodes as defined in \cite{Influenced_score}, where the influence score determines the local scope sensitivity of node $v$ on node $u$ as follows:

\begin{equation}
\label{eq_rgcn_regressor}
% \mathcal{L}
I(v,u) = \sum_{i}\sum_{j}\lvert\frac{\partial h_{ui}^{(L)}}{\partial X_{vj}}\rvert
\end{equation}

The equation $\eqref{eq_rgcn_regressor}$ represents the influence score between nodes $v$ and $u$ in a neural network model. In this equation, $h_{ui}^{(L)}$ refers to the $i$-th element in the embedding of node $u$ at the last layer $L$, while $X_{vj}$ represents the $j$-th feature of node $v$.
The partial derivative in equation $\eqref{eq_rgcn_regressor}$  measures the contribution of node $v$ to the embedding of node $u$ in the final layer $L$. A higher value indicates that the update from node $v$ has a more significant impact on node $u$.
This influence measurement is crucial for filtering out irrelevant neighbour nodes in a given task. The influence score function can be implemented using various methods, such as a weighted random walk function with $l$ steps or the shortest path distance. 
IBS potentially increases neighbour node type entropy by strategically 
selecting nodes based on their influence and ensures reachability by expanding from target vertices to the most influencing neighbour nodes.\shorten

% \textcolor{blue}{IBS regularize the URW subgraph with $\mathcal{V}_T$ as starting node set and PPR while expanding to neighbour nodes that increase the sampled subgraph neighbour-node-type’s entropy. }
% \textcolor{blue}{The PPR focuses and biases the URW exploration strategy that allows the diverse node types connected with the task target nodes to be sampled which increases the neighbour types entropy. }

% These methods allow us to formalize the influence score according to the requirements of the application domain.
% This influence measurement is crucial for filtering out irrelevant neighbour nodes in a given task. Depending on the specific application domain, the influence score function can be implemented using various methods, such as a weighted random walk function with $l$ steps or the shortest path distance from a single source in traffic networks. These methods allow us to formalize the influence score according to the requirements of the application domain.

\vspace*{-0ex}
\begin{algorithm}[t]	\caption{Influence-based Sampling}
	\label{alg:IBS}
        % \algsetup{linenosize=\small}
        \small
	\begin{algorithmic}[1]
	    \Input \textit{Knowledge graph KG; GML task $\mathcal{A}$;batch size $bs$;  top-k $k$;}
	   \Output Knowledge graph \(KG^\prime\).
	    \Function{IBS}{KG, $\mathcal{A}$, $pn$, $k$}:
	  \State Inf-mat ${\leftarrow} \text{getInfluenceScore} (KG, \mathcal{A}.\mathcal{V}_T)$ 
       \State $topk\_pairs \leftarrow \text{SelectTopK-Nodes} (\mathcal{A}.\mathcal{V}_T,\text{Inf-mat},k)$ 
       \State $gPartition \leftarrow \text{getPartition} (KG,topk\_pairs,bs)$
       \State $KG^{\prime} \leftarrow \text{extractSubgraph} (KG,gPartition)$
	\EndFunction
	\end{algorithmic} 
\end{algorithm}

% where $h_{ui}^{(L)}$ is i'th element in the embedding of node $u$ in the last neural network layer $L$ and  $X_{vj}$ is the feature $j$ of node $v$. This partial derivative simply measures the update of the node $v$ contributed to node $u$ embedding at the final layer $L$. The higher the update the more important the node $v$ to node $u$.
% Measuring the influence (importance) of node $v$ on target node $v_t$ is essential to filter out the task-irrelevant neighbour nodes. The influence score function can be implemented as a weighted random walk function with $l$ steps and be formalized using different methods according to the application domain,i.e., single source shortest path distance in traffic networks.

Algorithm \ref{alg:IBS} is the pseudocode of IBS. The influence score between each target node in $\mathcal{V}_T$ and all neighbour nodes is calculated using the approximate PPR in lines 2-3 on a homogeneous graph. The top-k pairs of target neighbour nodes are selected to construct a graph partition of $bs$ target nodes in line 4. \highlightedReply{The functions at lines 2 to 4 are parallelized using multi-threading.}{R2.O3} 
The graph partition contains the neighbour nodes with high overlap between the $bs$ nodes for efficient batches. The graph partition nodes and their in-between edges in the original KG are used to construct the $KG^{\prime}$ subgraphs. 
The IBS sampler preserves the local graph structure of the task by selecting nodes connected to $\mathcal{V}_T$ with the highest influence scores. Simultaneously, it takes into account all edges among the chosen nodes to construct $KG^{\prime}$, thereby preserving the global graph structure of the task.
After the graph partition selection in line 4, we transform back the homogeneous subgraph into a heterogeneous subgraph by adding the node/edge types. The GCN layers are replaced with RGCN layers to learn the heterogeneous graph semantics effectively.\shorten

Two factors dominate the Complexity of IBS. Firstly the PPR to expand for neighbour nodes with complexity $\mathcal{O}(\frac{N_{out}}{\mathcal{\epsilon}\mathcal{\alpha}})$ where $\mathcal{\epsilon}$ is the error value that goes to zero and $\mathcal{\alpha}$ is the PPR teleport probability which is very small value in dense graphs (between 0.1-0.25). This complexity is dependent on the number of non-target nodes and the graph density. Secondly, the induced subgraph extraction in line 5 with complexity  $\mathcal{O}(d|\mathcal{V}_s|)$, where $d$ is the average degree of nodes in $gPartition$ that depends on the structure of the graph. The large $k$ and $bs$ lead to a large subgraph size that requires larger training memory and time. Tuning these parameters helps achieve good balance based on the graph structure and density. However, the user should be aware of these factors for a given KG. 
Our IBS samples the neighbouring nodes based on their importance. Hence, it may lead to a better model's performance. However, its complexity will lead to excessive computation to extract the TOSG from a dense KG with a large number of vertices. Extracting the TOSG aims to reduce the overall training time and memory usage. Hence, it is necessary to optimize the cost of extracting the TOSG.\shorten

%Influence score calculation in Section \ref{sec:ibs} used the PPR to decide the important neighbour nodes that introduced task-relevant subgraphs but it comes with a high cost of PPR especially in a large dense graph and tasks with a high ratio of target nodes.

\vspace{-1ex}
\subsection{A SPARQL-based Extraction Method}
\label{sec:kg-tosa}

The SPARQL-based method is designed to offload the extraction of subgraphs matching our generic graph pattern to an RDF engine. This facilitates the integration of GNN training systems onto existing RDF engines, which are widely used to store KGs~\cite{KGNet,SPARQL_ML}. Our generic graph pattern for $KG'$, as illustrated in Figure~\ref{fig:node-BGP-samplers}, could be formalized as a basic graph pattern (BGP), which could be performed as a SPARQL query. Hence, the SPARQL-based method efficiently leverages the built-in indices, which RDF engines construct by default. Moreover, unlike BRW and IBS, the SPARQL-based method eliminates the need for a complete migration of the KG outside the RDF engine. This full migration proves to be computationally expensive with large KGs~\cite{KGNet,SPARQL_ML}.
The SPARQL-based method preserves: \myNum{i} the local graph structure by considering connected nodes to $\mathcal{V}_T$, and \myNum{ii} the global structure by merging the generated triples to construct $KG^{\prime}$.

    \vspace*{1ex}
\begin{algorithm}[t]
\caption{SPARQL-based TOSG Extraction}
	\label{alg:SPARQL_MS}
        % \algsetup{linenosize=\small}
        \small
	\begin{algorithmic}[1]
	    \Input \textit{KG ID $KG_{ID}$; GML task $\mathcal{A}$; Number of Direction $d$ hops $h$; SPARQL Endpoint $SP$; batch size: $bs$, number of threads: $\mathcal{P}$}
	    \Output Knowledge graph \(KG'\).
	    \Function{SPARQL\_MS}{$KG_{ID}$, $\mathcal{A}$, $d$, $h$, $SP$,$bs$,$\mathcal{P}$}:
      \State$ BGP \leftarrow \textrm{getBGP(}\mathcal{A},d,h).$ 
      \State$ count \leftarrow getGraphSize(KG_{ID}, SP, BGP)$
      \State$ \mathcal{QB}\leftarrow executionPlanner(KG_{ID}, SP, BGP, count, bs)$
      \State$index = 0$ and initializeWorkers($\mathcal{P}$) 
      \State\Worker[RequestHandler]:
      \While{$i < \mathcal{QB}.size()$} \Comment{loop over the query batches}
        \State$ \mathcal{T}riples \textrm{ +}= sparqlExec(KG_{ID}, SP, \mathcal{QB}[index\textrm{++}])$ 
      \EndWhile 
      \State$ KG^\prime_{csv} \leftarrow dropDuplicates(\mathcal{T}riples)$ 
	\EndFunction
	\end{algorithmic}
 \end{algorithm} 

    % \State$ KG^\prime_{csv-r} \leftarrow SparqlExecution(KG_{ID}, SP, \mathcal{QB}, \mathcal{P})$

Our SPARQL-based method could be tuned to generate query variations by adjusting two main parameters: \myNum{i} predicate direction $d$ and \myNum{ii} the number of hops $h$. These parameters control the average distance of nodes from the target nodes and incorporate various node/edge types. In {\sysName}, the BGP query will include by default the outgoing predicates, i.e., $d = 1$, and one hop, $h = 1$ ({\sysName}$_{d1h1}$). TOSG can include three more variations: \myNum{i} ({\sysName}$_{d2h1}$) outgoing and incoming predicates, i.e., $d = 2$, and one hop, $h = 1$,  \myNum{ii} ({\sysName}$_{d1h2}$) outgoing predicates, i.e., $d = 1$, and two hop, $h = 2$, and \myNum{iii} ({\sysName}$_{d2h2}$) outgoing and incoming predicates, i.e., $d = 2$, and two hop, $h = 2$. A SPARQL query ($Q^{d2h1}$) for {\sysName}$_{d2h1}$ could be formulated as:

\begin{figure}[h]
\vspace*{-3ex}
  \centering    
  \lstdefinestyle{myStyle}{basicstyle=\small\ttfamily, language=SPARQL }
\begin{lstlisting}    [style=myStyle,captionpos=b,showspaces=false,numbers=left,xleftmargin=0.5cm,morekeywords={v,a,o} ,escapechar=\%,showstringspaces=false,columns=fullflexible]
select ?s ?p ?o {
    select ?v as ?s ?p ?o 
    where { ?v a <Node_Type_URI>.
            ?v ?p ?o.}
    union select ?s ?p ?v as ?o 
    where  {?v a <Node_Type_URI>.
            ?s ?p ?v.} }
\end{lstlisting}
  % \caption{A SPARQL-based query for TCGP with d=2 and h=1.}
  % \label{fig:SPARQL-BGP}
  \vspace*{-3ex}
\end{figure}

{\sysName} supports all these variations.  
These four variations are used for node classification tasks. 
In the case of a link prediction task targeting vertices of two different types, we add the triple pattern \RDFTYPE{?$v_{Ti}$}{$p_T$}{?$v_{Tj}$} between the two subgraphs extracted for ?$v_{Ti}$ and ?$v_{Tj}$. \highlightedReply{Algorithm~\ref{alg:SPARQL_MS} is the 
pseudocode of our parallel SPARQL-based TOSG extraction.}{R2.O3} It formalizes a BGP for a given task based on a certain $d$ direction and $h$ hops in line 2. However, this query may target a large number of vertices, which can cause issues, e.g., network congestion or low bandwidth. To address this, {\sysName} uses compression and pagination optimization techniques when dealing with query results of a large number of triples. Most RDF engines support compression. Hence, {\sysName} sends an HTTP request to the SPARQL endpoint with a compression flag.\shorten

Existing RDF engines support pagination using LIMIT and OFFSET. 
By dividing the results into $k$ mini-batches, most RDF engines will execute the query $k$ times. This is inefficient as our queries are formulated with UNION, such as line 5 in $Q^{d1h1}$. Repeating the UNION query $k$ time is time-consuming due to duplicate elimination. Thus, Algorithm~\ref{alg:SPARQL_MS} paginates each subquery independently. 
Each subquery will benefit from the RDF built-in indices, as the query targets a vertex of a known type.
Existing RDF engines support six indexing schemes for traditional lookup on any subject, predicate or object in a SPARQL query~\cite{Hexastore}.
% ~\cite{Hexastore, TripleBit}
Thus, our SPARQL-based queries are executed efficiently by leveraging the indices existing in RDF engines.
Algorithm~\ref{alg:SPARQL_MS} collects in $\mathcal{P}$ parallel threads the set of triples in a Pandas DataFrame (DF) and finally uses DF to eliminate duplicates. The algorithm gets as input the batch size per HTTP request. These optimizations are performed in lines 3 to 10.
The complexity of {\sysName} is dominated by $\mathcal{QB}$ size in line 7 and the average execution time per query. Also, the drop duplicates step at line 10 is performed in $\mathcal{O}(|KG^{\prime}|)$ which is dominated by the small size of $|KG^{\prime}|$. The $d$ and $h$ parameters contribute to the average query execution time and as well the size of $KG^{\prime}$.

The merging process at line 8 enables {\sysName} to maintain longer metapaths. For example, for the metapath Author-Writes-Paper-Cites-Paper-PublishedIn-Venue in MAG, the subgraphs of common vertices will be interconnected. This process forms a larger sub-graph (TOSG) with longer metapaths while still maintaining a smaller number of hops ($h$) from the target vertices (in this case, vertices of type "Publication"). By using longer metapaths, the SPARQL-based method enables HGNNs to achieve better semantic attention. It also allows for constructing multi-layered GCNs with a limited number of hops. Compared to BRW and IBS, the SPARQL-based method has lower complexity since it does not require sampling on the entire graph and utilizes the RDF engine's built-in indices. Hence, 
The SPARQL-based method ensures a decrease in both training time and memory usage while extracting a TOSG of comparable graph quality to those obtained by BRW and IBS in terms of data sufficiency and graph topology.\shorten

% the SPARQL-based method guarantees a reduction in the overall training time and memory usage while extracting a TOSG of comparable graph quality to the one extracted by BRW and IBS in terms of data sufficiency and graph topology.\shorten 

% \input{sections/benchmark.tex}
\section{EXPERIMENTAL EVALUATION}
\label{sec:expermients} 

\subsection{Evaluation Setup}

\begin{table}[t]
\vspace*{-1ex}
\centering
\caption{Our Benchmark Statistics. The number of nodes and edges (RDF triples) is in millions. The number of node types (n-type) and edge types (e-type) is tens to thousands.\shorten
}
\vspace*{0ex}
\label{tbl_exp_ds_Statistics}
\begin{tabular}{lrrrrr}\hline
KG-Dataset &\#nodes &\#edges &\#n-type &\#e-type \\\midrule
MAG-42M &42.4M &166M &58 &62 \\
YAGO-30M &30.7M &400M &104 &98 \\
DBLP-15M &15.6M &252M &42 &48 \\ %\midrule
% FB15K-237 &14.5K &310K &- &237 \\
% WN18 &40.9K &92K &- &11 \\
ogbl-wikikg2&2.5M&17M&9.3K&535 \\
YAGO3-10 &123K &1.1M &23&37 \\
% OBGN-MAG&1.9M&62M&4&4\\
% DBLP-26K&26K&119K&4&3\\
% IMDB-12K&12K&18K&3&2\\
\hline
\end{tabular}
\vspace*{-3ex}
\end{table}

\begin{table}[t]
\vspace*{-1ex}
\centering
\caption{A summary of our GNN tasks: Task Types (TT) are single-label node classification (NC) or missing entity link prediction (LP).\shorten}
\vspace*{0ex}
\label{tbl_exp_tasks}
\begin{tabular}{cccccc}\hline
TT&Name&KG&Split&Ratio&Metric \\\midrule
NC &PV&MAG-42M&Time&84/9/7&Accuracy \\
NC &PD&MAG-42M&Time&87/8/5&Accuracy \\
NC &PC&YAGO-30M&Random&80/10/10&Accuracy\\
NC &CG&YAGO-30M&Random&80/10/10&Accuracy\\
% Person-Profession/LP &FB15K-237 &Profession &- &Random &MRR-Hits@10 \\
% Word Hyponym/LP &WN18 &Hyponym &- &Random &MRR-Hits@10 \\
NC &PV&DBLP-15M&Time&79/10/11&Accuracy\\
NC &AC&DBLP-15M&Time&80/10/10&Accuracy \\\midrule
LP &AA&DBLP-15M&Time&99/0.7/0.3&Hits@10\\
LP &PO&ogbl-wikikg2&Time&94/2.5/3.5&Hits@10\\
LP &CA&YAGO3-10&Random&99/0.5/0.5&Hits@10\\ \midrule

% NC&OBGN-MAG&\textbf{Paper}/PublishedIn/Venue\\
% NC&DBLP-26K&\textbf{Author}/Discipline/FieldOfResearch\\
% NC&IMDB-12K&\textbf{Movie}/Type/Genres\\
% \bottomrule
\end{tabular}
\vspace*{-3ex}
\end{table}

% \begin{table*}[ht]\centering
% \centering
% \caption{The benchmark datasets statistics.}
% \label{tbl_exp_ds_Statistics}
% \scriptsize
% \begin{tabular}{|p{11em}|p{6em}|p{5em}|p{3em}|p{6em}|p{6em}|p{5em}|p{6em}|}
% % p{6em}|p{10em}|}
% \toprule
% Dataset & Abbreviation & KG & Task & \#FM Triples  & \# RS Triples  & \# SQ Triples & \# Node Types
% % &\# Edge Types& \# graph density  
% \\
% \midrule
% MAG Paper-Venue & MAG\_PV&MAG& NC  &NAN &21M &NA&NA
% % && 
% \\
% % \midrule
% MAG Paper-Discipline & MAG\_PD&MAG& NC&NAN& 21M&NA&NA
% % && 
% \\
% % \midrule
% DBLP Paper-Venue & DBLP\_PV &DBLP& NC&252M & 64M  &NA&NA
% % && 
% \\
% % \midrule
% DBLP Author-Country & DBLP\_AC&DBLP& NC &252M & 6M  &NA&NA
% % && 
% \\
% % \midrule
% Yago4 Place-Country & YAGO\_PC &Yago& NC&250M & NA  &NA&NA
% % && 
% \\
% % \midrule
% Yago4 Work-Category & YAGO\_WC&Yago& NC &250M & NA  &NA&NA
% % && 
% \\
% \midrule
% FB Person-Profession&FB15k\_PP&Free-Base& LP &237K& -  &NA&NA
% % && 
% \\
% % \midrule
% Yago10 Connected-Airport&YAGO10\_CA&Yago& LP &1.1M& -  &NA&NA
% % && 
% \\
% % \midrule
% WN Word-Hyponym&WN18\_WH&Word Net& LP &142K& -  &NA&NA
% % && 
% \\
% DBLP Author-Affiliation&DBLP\_AA&DBLP& LP &252M& -  &NA&NA
% % && 
% \\
% % Wiki Person-Occupations& Wikikg\_PO & Wikikg-v2 & LP &16.5M & NA &NA&NA& &\\
% \bottomrule
% \end{tabular}
% \end{table*}

\begin{figure*}[t]
\vspace*{-2ex}
     \centering
     % \hfill
     \begin{subfigure}[b]{0.99\textwidth}
     \vspace*{-1.7ex}
         \centering         \includegraphics[width=0.99\textwidth]{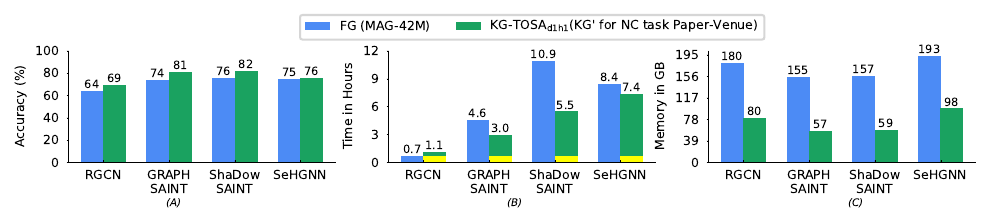}
         % \caption{MAG-42M (Paper-Venue)}
         \label{fig:MAG-42M-RW}
     \end{subfigure}
     % \hfill
     \begin{subfigure}[b]{0.99\textwidth}
      \vspace*{-1.7ex}
         \centering         \includegraphics[width=0.99\textwidth]{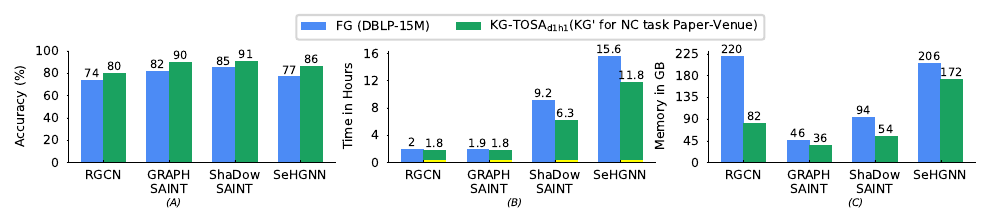}        
         % \caption{DBLP-15M (Paper-Venue)}
         \label{fig:DBLP-15M-RW}
     \end{subfigure}
     \begin{subfigure}[b]{0.99\textwidth}
     \vspace*{-1.7ex}
         \centering        \includegraphics[width=0.99\textwidth]{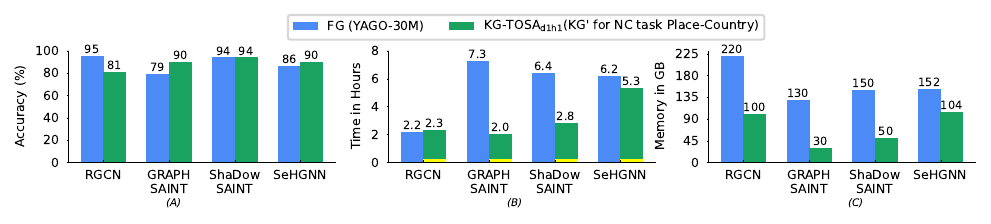}
         % \caption{YAGO-30M (Place-Country)}
         \label{fig:YAGO-30M-RW}
     \end{subfigure}
     \vspace*{-3ex}
    \caption{ 
    Performance across NC tasks is based on three metrics: (A) Accuracy (higher is better), (B) Training-Time (lower is better), and (C) Memory (lower is better). The top and middle sections illustrate the results for the Paper-Venue task on MAG and DBLP, respectively. The bottom figures present the outcomes for the place-country task. With {\sysName}, all methods experience improvements in accuracy in most cases while reducing memory and time, even with {\sysName}'s preprocessing time in yellow.\shorten 
    % The KG' extraction and transformation process by {\sysName} is highlighted in yellow.\shorten
    % Performance of RGCN, GraphSAINT, ShaDowSAINT, and SeHGNN in the NC tasks. (A) Accuracy (higher is better), (B) Training-Time (lower is better), (C) Training-Memory (lower is better). The figures on top and in the middle show the results for the paper-venue classification task on MAG and DBLP, respectively. The figures at the bottom show the results for the place-country classification task. {\sysName} enables all methods to reduce memory and time while improving accuracy in most cases.\shorten
    } 
\label{fig:KGTOSA_NC}
  \vspace*{-3ex}
\end{figure*}

\subsubsection{Our Benchmark (KG datasets and GNN tasks)}
Our KG datasets include real KGs, such as MAG, DBLP, YAGO and WikiKG. The datasets are extracted from real KGs of different application domains: general-fact KGs 
% (Yago~\footnote{YAGO-4: \url{https://yago-knowledge.org/downloads/yago-4}} and ogbl-wikikg2~\footnote{ogbl-wikikg2: \url{https://ogb.stanford.edu/docs/linkprop/\#ogbl-wikikg2}}) and academic KGs (MAG\footnote{MAG-2020-05-29: \url{https://makg.org/rdf-dumps/}} and DBLP~\footnote{DBLP versions March 2022 for NC and (Jan, March, and May) 2023 for LP: \url{https://dblp.org/rdf/release/}}). 
(YAGO4 and ogbl-wikikg2) , and academic KGs (MAG and DBLP). 
These KGs contain up to 42.4 million vertices, 400 million triples, and tens to thousands of node/edge types. 
Existing HGNN methods for LP tasks on large KGs require excessive computing resources. Hence, we also used YAGO3-10
% ~\footnote{YAGO3-10: \url{https://paperswithcode.com/dataset/yago}}
, a smaller version of YAGO, and ogbl-wikikg2~\cite{OGB}, a dataset extracted from Wikidata. It contains 2.5M entities and 535 edge types.
Tables~\ref{tbl_exp_ds_Statistics} and~\ref{tbl_exp_tasks} summarize the details of the used KGs and our defined NC and LP tasks~\footnote{Detailed explanation for our benchmark datasets/tasks are available \href{https://github.com/CoDS-GCS/KGTOSA/blob/main/KGTOSA_SupplementalMaterial.pdf}
{\textcolor{blue}{here}}.}, respectively. 
Our benchmark will be available for further study.\shorten

Our benchmark includes NC and LP tasks. In KGs, an NC task is categorized into single- or multi-label classifications, as defined in~\ref{def:NC}. Our benchmark followed existing datasets in ~\cite{OGB,SeHGNN,Shadow-GNN} and defined single-label NC tasks. We choose the accuracy metric for these NC tasks to evaluate the performance. The train-valid-test splits are either split using a logical predicate that depends on the task or stratified random split with 80\% for training, 10\% for validation and 10\% for testing. Table~\ref{tbl_exp_tasks} summarizes the details of the split schema and ratio. Our benchmark includes two NC tasks per KG.\shorten

For LP tasks, KGs contain many edge types that vary in importance to a specific task. For example, a GNN task predicting the academic affiliation will benefit from the predicates (edge type) \textit{discipline} and \textit{affiliation} in Wikidata but not from predicates related to \textit{movies} and \textit{films}. Moreover, training HGNNs for an LP task on a large KG of many edge types is computationally expensive. Hence, we define our LP tasks for a specific predicate to enable the HGNN methods to train the models with our available computing resources: a Linux machine with 32 cores and 3TB RAM. 
We choose the Hits@10 metric to evaluate the predictions ranking performance following SOTA methods~\cite{MorsE,OGB,Shadow-GNN}. The train-valid-test splits are either split using three versions of the KG based on time or randomly. The details of the LP tasks are summarized in Table~\ref{tbl_exp_tasks}.

\subsubsection{Computing Infrastructure} 
Our experiments used two different settings: \myNum{i} VM$^{GB}$: two Linux machines, each with dual 64-core Intel Xeon 2.4 GHz (Skylake, IBRS) CPUs and 256GB RAM, and \myNum{ii} VM$^{TB}$: one Linux machine with 32 cores and 3TB RAM. 
%
% {\sysName} uses standard SPARQL queries to support different RDF engines. Virtuoso is used as a SPARQL Endpoint for several real-world KGs, such as YAGO4, DBLP and MAG. Thus, 
We use the standard, unmodified installation of Virtuoso 07.20.3229. On VM$^{GB}$, we installed one Virtuoso instance per KG. We run each method in VM$^{TB}$ for the NC task. We use VM$^{GB}$ for the LP task on Yago3-10 and VM$^{TB}$ for the LP task on DBLP and ogbl-wikikg2.\shorten

\subsubsection{Compared GNN Methods} 
\highlightedReply{We evaluate {\sysName} performance against our developed baselines BRW and IBS using the state-of-the-art (SOTA) GNN methods as follows:}{R2.O1} \myNum{i} RGCN-based methods, such as ShaDowSAINT~\cite{Shadow-GNN} and GraphSAINT\cite{GraphSAINT}, a metapath-based method, namely SeHGNN~\cite{SeHGNN}. These SOTA methods are based on sampling and mini-batch training, \myNum{ii} RGCN\footnote{{ \href{https://github.com/snap-stanford/ogb/blob/master/examples/nodeproppred/mag/rgcn.py}{\textcolor{blue}{RGCN+}} implementation used for NC and 
\href{https://github.com/pyg-team/pytorch_geometric/blob/master/examples/rgcn_link_pred.py}{\textcolor{blue}{RGCN-PYG}}
implementation used for LP.}}~\cite{RGCN}, full-batch training (no sampling), and \myNum{iii} MorsE~\cite{MorsE} and LHGNN \cite{LHGNN}. We used the code available in the GitHub repositories provided by the authors of each method. The code of SeHGNN, ShaDowSAINT, and GraphSAINT support only node classification tasks. MorsE and LHGNN are the SOTA methods in link prediction, and their code does not support node classification tasks. The RGCN code supports both tasks. \highlightedReply{We used the default tuned training parameters provided by the authors of each method as indicated in their available code and paper and initialized node embeddings randomly using Xavier weight.}{R3.O1} MorsE-TransE version is used in the LP tasks.
{\sysName} utilizes by default the SPARQL-based method to extract $KG'$ from $KG$ for a specific task. Then, each HGNN method is tested twice: \myNum{i} using the original $KG$ denoted as a full graph (FG), and \myNum{ii} using the $KG'$. We perform the three runs of each method on the exact VM and report the metric average value.\shorten

\subsection{The {\sysName} Impact on HGNN Methods}
We evaluate the performance of each method in terms of the model performance, training time, and memory consumption for each task. Below, we present our results by task type. {\sysName}$_{dihj}$ has two primary parameters: $d$ (1 or 2) and $h$ (number of hops). For NC tasks, we utilized {\sysName} with only outgoing predicates ($d = 1$) and one hop ($h = 1$), which we refer to as {\sysName}$_{d1h1}$. For LP tasks, we utilized {\sysName} with bidirectional predicates ($d = 2$) and one hop ($h = 1$), which we refer to as {\sysName}$_{d2h1}$.\shorten

\vspace*{0ex}
\subsubsection{Node Classification Tasks}
\label{subsec_CTasks}
We utilized {\sysName}$_{d1h1}$ to extract $KG'$ for each NC task/KG. Figure~\ref{fig:KGTOSA_NC} presents the performance of each GNN method using the full KG (FG) and $KG'$. The preprocessing overhead associated with extracting and transforming the $KG'$, denoted in yellow in Figure~\ref{fig:KGTOSA_NC}.(B), is negligible compared to the overall savings. Due to space constraints, we only show the results for three NC tasks~\footnote{The remaining tasks are available in the supplementary materials
% \href{https://gitfront.io/r/HGNN/LWYY16iUMSuE/KGTOSA/raw/KGTOSA_SupplementalMaterial.pdf}
\href{https://github.com/CoDS-GCS/KGTOSA/blob/main/KGTOSA_SupplementalMaterial.pdf}
{\textcolor{blue}{here}}.}. In general, {\sysName} enables all SOTA methods for the six NC tasks to decrease training time and memory usage while improving accuracy in most cases. The improvement achieved by {\sysName} varies based on different factors, such as the sampling support and the size of $KG'$ compared to FG. The latter depends on the diversity of FG and the number of target vertices in a task. For instance, MAG-42M and YAGO-30M are diverse KGs with around 60 to 100 node/edge types, while YAGO-30M and DBLP-15M are the largest in terms of triples, with 400M and 252M, respectively, as reported in Table~\ref{tbl_exp_ds_Statistics}.

\begin{figure}[t]
% \vspace*{-1ex}
     \centering
     % \hfill
     \begin{subfigure}[b]{0.99\columnwidth}
     \vspace*{-4ex}
         \centering         \includegraphics[width=\textwidth]{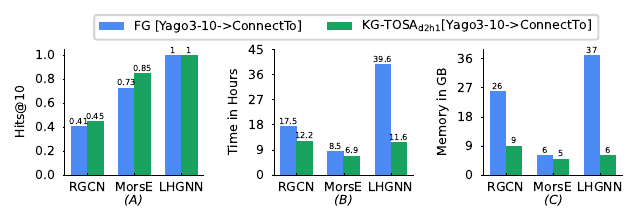}
         % \caption{MAG-42M (Paper-Venue)}
         \label{fig:LP_YAGO10}
     \end{subfigure}
       % \hfill
     \begin{subfigure}[b]{0.99\columnwidth}
     \vspace*{-4ex}
         \centering         \includegraphics[width=\textwidth]{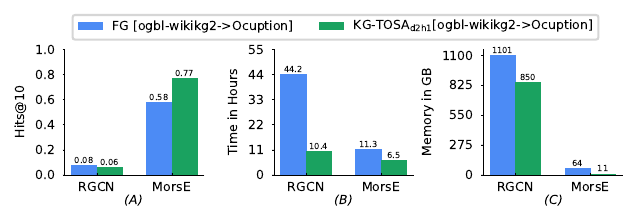}
         % \caption{MAG-42M (Paper-Venue)}
         \label{fig:LP_Wikikg}
     \end{subfigure}
     % \hfill
     \begin{subfigure}[b]{0.99\columnwidth}
      \vspace*{-4ex}
         \centering         \includegraphics[width=\textwidth]{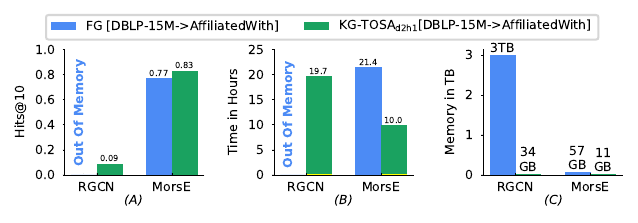}        
         % \caption{DBLP-15M (Paper-Venue)}
         \label{fig:LP_DBLP}
     \end{subfigure}
     \vspace*{-7ex}
    \caption{ 
Performance of RGCN, MorsE, and LHGNN in the LP tasks. 
  (A) Hits@10 (higher is better), (B) Training-Time, (C) Training-Memory.
  RGCN exceeded the 3TB RAM in VM$^{TB}$.
  % {\color{blue} RGCN experiments are still runing}
  } 
\label{fig:KGTOSA_LP}
\vspace*{-3ex}
\end{figure}

RGCN is a full-batch GNN method without performing any sampling unlike other methods, such as GraphSAINT, ShaDowSAINT, and SeHGNN. Hence, RGCN has the shortest training time, but it consumes excessive memory. As a result, {\sysName} enables RGCN to significantly decrease its memory consumption, up to 63\%, with DBLP-15M for the paper-venue task. RGCN is the method that benefits the least from {\sysName} in terms of training time. However, GraphSAINT, ShaDowSAINT, and SeHGNN significantly benefit from the extracted $KG’$ by {\sysName}, which empowers them to reduce their time and memory usage across all tasks and KGs with improving in accuracy by up to 11\%. 
% RGCN on $KG'$  achieves huge memory savings compared to the FG on the sampling methods.\shorten

% Furthermore, {\sysName} enables them to improve accuracy by up to 11\%.\shorten

% \begin{figure*}[t]
% \vspace*{-3ex}
%   \centering  
%   \includegraphics [width=\textwidth]{figures/KGTOSA_LP_MRR_Time_Memory_Results.pdf}
%   \vspace*{-7ex}
%   \caption{Performance of RGCN and MorsE in the LP tasks. 
%   (A) Mean reciprocal rank  (MRR Hits@10 higher is better), (B) Training-Time, (C) Training-Memory.
%   RGCN is out of memory in a VM with 3TB RAM. Unlike NC, LP tasks consume more memory for generating positive and negative samples. {\sysName} achieves an outstanding memory reduction for RGCN and MorsE.\shorten}  
%   \label{fig:KGTOSA_LP}
% \end{figure*}

% \begin{figure*}[t]
% \vspace*{-3ex}
%   \centering
%    \includegraphics [width=\textwidth]{figures/KG_TOSA_VS_Biased_Acc_Time_Memory_Results.pdf}
%    \vspace*{-6ex}
%   \caption{ FG Accuracy using GraphSAINT BRW sampler v.s. {\sysName} meta-samplers. {\sysName} meta-samplers achieve significant improvement in training time and memory with comparable to better accuracy. }
%   \label{fig:KG_TOSA_VS_Biased_Acc_Time_Memory_Results}
% \end{figure*}

\subsubsection{Link Prediction Tasks}
\label{subsec_LPTasks}
We utilized {\sysName}$_{d2h1}$ to extract $KG'$ for each LP) task. Figure~\ref{fig:KGTOSA_LP} shows the performance of RGCN, MorsE and LHGNN using the full KG (FG) and our $KG'$. The preprocessing overhead (denoted in yellow) of {\sysName}$_{d2h1}$ is reported in Figure~\ref{fig:KGTOSA_LP}.(B). This preprocessing is very small w.r.t the training time. Hence, it is hard to be recognized. For the DBLP LP task, our preprocessing is 14 minutes, while training time is 10 hours. {\sysName} enables RGCN, LHGNN, and MorsE to reduce the training time and memory consumption while improving the model's performance (higher  Hits@10) in most cases, as shown in figure~\ref{fig:KGTOSA_LP}. LHGGN achieved the highest score and consumed excessive time and memory. Hence, LHGGN did not finish training for other LP tasks on the larger KGs, i.e., wikikg2 and DBLP-15MK.\shorten 

RGCN and MorsE consume more memory and time with large KGs, such as DBLP-15M, w.r.t smaller ones, e.g., YAGO3-10 and ogbl-wikikg2. RGCN failed to train the task on DBLP-15M using VM$^{TB}$ with 3TB RAM. However, {\sysName} enabled RGCN to finish this task with only 35GB RAM. This is tremendous memory savings of more than 98\%. Moreover, {\sysName} empowered MorsE to improve the accuracy of DBLP from 0.77 to 0.83 and ogbl-wikikg2 from 0.58 to 0.77. This improvement demonstrates that as tasks' complexity and the size of KGs increase, {\sysName}  enables GNN methods to achieve more performance improvements (better Hits@10) while reducing time and memory usage.

Our defined LP tasks are based on one given predicate. Performing LP for all edge types, i.e., KG completion, consumes excessive computing resources. Hence, training HGNNs on large KGs might not be feasible due to a lack of resources. For example, performing KG completion using MorsE on DBLP-15M consumed 330GB memory and 124 training hours compared with 11 GB and 9.8 training hours using the KG' of {\sysName} for \textit{affaliatedWith} edge type only. This huge saving in memory and time (one order of magnitude less time and memory) enables users to perform LP on predicates of their interest. Moreover, we can efficiently train LP tasks on a set of individual predicates in parallel.\shorten

\begin{table*}[t]
\centering
\vspace*{-3ex}
\caption{ Statistics of extracted subgraphs using URW, BRW, IBS, and {\sysName}$_{d_1h_1}$ on Yago-4, DBLP, and MAG KGs. \highlightedReply {The walk length h (no. of hops) is
= 3 and the size of the initial set of target nodes is 20k.}{R2.O2} Our SPARQL-based method using $_{d_1h_1}$ achieves similar statistics and comparable accuracy to BRW and IBS. However, BRW and IBS consume excessive computing resources w.r.t the SPARQL-based method as shown in Figure~\ref{fig:DifferentParameters}.}
\vspace*{0ex}
\label{tab:tbl_YAGO_compare_stat}
% \small
\scriptsize
\begin{tabular}{|c|cc|cc|cc|cc|cc|cc|cc|cc|c|}
\hline
\textbf{Indicator} & \multicolumn{4}{c|}{$|\textbf{KG}^{\prime}|$}& \multicolumn{12}{c|}{\textbf{Data Sufficiency}} \\ \hline
 \textbf{Task}& \multicolumn{4}{c|}{\textbf{$|\mathcal{V}|$}} & \multicolumn{4}{c|}{\textbf{$\mathcal{V}_T$ (\%)}} & \multicolumn{4}{c|}{\textbf{$|\mathcal{C}'|$}} & \multicolumn{4}{c|}{\textbf{$|\mathcal{R}'|$}}
\\\hline
&\multicolumn{1}{c|}{RW} & \multicolumn{1}{c|}{BRW}&\multicolumn{1}{c|}{IBS}&\multicolumn{1}{c|}{$d_1h_1$}&\multicolumn{1}{c|}{RW} & \multicolumn{1}{c|}{BRW}&\multicolumn{1}{c|}{IBS}&\multicolumn{1}{c|}{$d_1h_1$}&\multicolumn{1}{c|}{RW} & \multicolumn{1}{c|}{BRW}&\multicolumn{1}{c|}{IBS}&\multicolumn{1}{c|}{$d_1h_1$}&\multicolumn{1}{c|}{RW} & \multicolumn{1}{c|}{BRW}&\multicolumn{1}{c|}{IBS}&\multicolumn{1}{c|}{$d_1h_1$}
\\ \hline
CG/YAGO& \multicolumn{1}{c|}{55720}& 40936  & \multicolumn{1}{c|}{\textbf{78562}}&50429& \multicolumn{1}{c|}{1.1}&61.2& \multicolumn{1}{c|}{\textbf{82.4}}& 35.7& \multicolumn{1}{c|}{\textbf{69}}& \multicolumn{1}{c|}{36}& \multicolumn{1}{c|}{33}& \multicolumn{1}{c|}{34}&\multicolumn{1}{c|}{\textbf{64}}&\multicolumn{1}{c|}{31}&\multicolumn{1}{c|}{33}&    \multicolumn{1}{c|}{33}\\ \hline
PC/YAGO      & \multicolumn{1}{c|}{55299}    & 42016  & \multicolumn{1}{c|}{\textbf{79830}}          &50419& \multicolumn{1}{c|}{11.4}&55.3 & \multicolumn{1}{c|}{\textbf{74.6}}&35& \multicolumn{1}{c|}{\textbf{85}}&27&\multicolumn{1}{c|}{41}&25&\multicolumn{1}{c|}{\textbf{80}}&22& \multicolumn{1}{c|}{39}&\multicolumn{1}{c|}{24}\\ \hline
PV/DBLP  & \multicolumn{1}{c|}{40618}    & 44886  & \multicolumn{1}{c|}{\textbf{128026}}         & 51833      & \multicolumn{1}{c|}{29.9}& \textbf{65.4}& \multicolumn{1}{c|}{19.4}&40.2&\multicolumn{1}{c|}{\textbf{34}}&33& \multicolumn{1}{c|}{32}&24& \multicolumn{1}{c|}{33}&\textbf{34}& \multicolumn{1}{c|}{33}&\multicolumn{1}{c|}{25}\\ \hline
PV/MAG   & \multicolumn{1}{c|}{36149}     & 44956   & \multicolumn{1}{c|}{\textbf{92103}}          & 56744       & \multicolumn{1}{c|}{4.9}&\textbf{78}& \multicolumn{1}{c|}{26.5}& 36.2&\multicolumn{1}{c|}{\textbf{48}}& 20& \multicolumn{1}{c|}{23}& 26& \multicolumn{1}{c|}{\textbf{34} }   &22 & \multicolumn{1}{c|}{29}&\multicolumn{1}{c|}{25}\\ \hline
%%%%%%%%%%%%%%%%%%%%%%%%%%%%%%%%%%%
\textbf{Indicator} & \multicolumn{12}{c|}{\textbf{Graph Topology}}& \multicolumn{4}{c|}{\textbf{Other Statistics}} \\ \hline
 \textbf{Task}& \multicolumn{4}{c|}{\textbf{Target-Discon.(\%)}} & \multicolumn{4}{c|}{\textbf{Avg.Dist.Target}} & \multicolumn{4}{c|}{\textbf{Avg.Entropy(Eq.\ref{eq_shannon})}} & \multicolumn{4}{c|}{\textbf{Accuracy}}
\\\hline
&\multicolumn{1}{c|}{RW} & \multicolumn{1}{c|}{BRW}&\multicolumn{1}{c|}{IBS}&\multicolumn{1}{c|}{$d_1h_1$}&\multicolumn{1}{c|}{RW} & \multicolumn{1}{c|}{BRW}&\multicolumn{1}{c|}{IBS}&\multicolumn{1}{c|}{$d_1h_1$}&\multicolumn{1}{c|}{RW} & \multicolumn{1}{c|}{BRW}&\multicolumn{1}{c|}{IBS}&\multicolumn{1}{c|}{$d_1h_1$}&\multicolumn{1}{c|}{RW} & \multicolumn{1}{c|}{BRW}&\multicolumn{1}{c|}{IBS}&\multicolumn{1}{c|}{$d_1h_1$}
\\ \hline

CG/YAGO& \multicolumn{1}{c|}{\textbf{76.7}}    & 0  & 
\multicolumn{1}{c|}{0}&0&
\multicolumn{1}{c|}{\textbf{7.1}}& 4.23 & 
\multicolumn{1}{c|}{4.7}& 4.18& 
\multicolumn{1}{c|}{1.27}&2.68& 
\multicolumn{1}{c|}{\textbf{3.02}}&2.34&
\multicolumn{1}{c|}{15.25}&36.73 & \multicolumn{1}{c|}{\textbf{42}}&    \multicolumn{1}{c|}{36.72} \\ \hline
PC/YAGO & \multicolumn{1}{c|}{\textbf{14.8}}&0& \multicolumn{1}{c|}{0}&0&
\multicolumn{1}{c|}{\textbf{7.46}}&4.12& 
\multicolumn{1}{c|}{5.2}&4.62& 
\multicolumn{1}{c|}{1.27}&2.67& 
\multicolumn{1}{c|}{\textbf{2.96}}&2.40&
\multicolumn{1}{c|}{79.28}&96.1 & \multicolumn{1}{c|}{\textbf{97.2}}&    \multicolumn{1}{c|}{89.52} \\ \hline
PV/DBLP  & \multicolumn{1}{c|}{0}& 0 & 
\multicolumn{1}{c|}{0}&0&
\multicolumn{1}{c|}{\textbf{4.23}}&3.71& 
\multicolumn{1}{c|}{3.95}&3.1& 
\multicolumn{1}{c|}{1.77}&\textbf{2.75}& 
\multicolumn{1}{c|}{1.64}&2.18&
\multicolumn{1}{c|}{81.79}&80.53 & \multicolumn{1}{c|}{85.4}&    \multicolumn{1}{c|}{\textbf{89.52}} \\ \hline
PV/MAG   & \multicolumn{1}{c|}{\textbf{89.3}} &0& 
\multicolumn{1}{c|}{0}&0&
\multicolumn{1}{c|}{3.1}& 2.9& 
\multicolumn{1}{c|}{\textbf{3.2}}& 3.00& 
\multicolumn{1}{c|}{1.49}&\textbf{4.44}& 
\multicolumn{1}{c|}{2.36}&3.18&    \multicolumn{1}{c|}{73.79}&75.33&
\multicolumn{1}{c|}{75.4}&\textbf{81.08} \\ \hline
\end{tabular}
% \vspace*{-3ex}
\end{table*}

\subsection{Analyzing {\sysName}}

\subsubsection{Analyzing our Extraction Methods in Overall Performance}
\begin{figure}[t]
\vspace*{-3ex}
  \centering
   \includegraphics [width=\columnwidth]{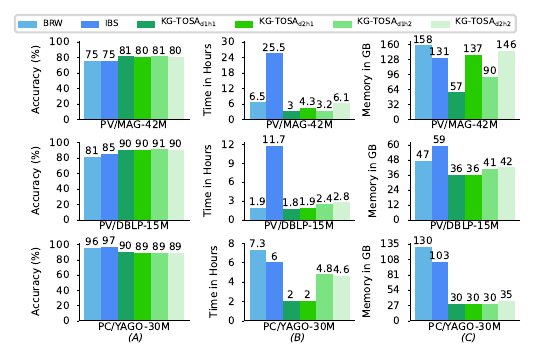}
   \vspace*{-4ex}
  \caption{ (A) Accuracy, (B) Extraction, Transformation, and Training Time, (C) Memory. GraphSAINT+BRW on FG v.s. Our SPARQL-based method with different parameters ({\sysName}$_{dihj}$). PV/MAG-42M at the top, PV/DBLP-15M in the middle, and PC/YAGO-30M at the bottom. {\sysName}$_{dihj}$ achieves comparable accuracy to BRW and IB while enabling HGNN methods to reduce training time and memory usage.\shorten}
  \label{fig:DifferentParameters}
  \vspace*{-3ex}
\end{figure}

\highlightedReply{These experiments compare our SPARQL-based method against our developed baseline methods (BRW and IBS)}{R2.O1} for extracting the TOSG. Our BRW implementation adopts the GraphSAINT subgraph sampler~\cite{GraphSAINT}. We replaced the RW subgraph sampler with our implemented BRW sampler. 
\highlightedReply{The parameters of BRW are the same for all tasks: $bs=20000$, $h=3$ and RGCN embedding-dim=128.
For IBS, the parameters are the same for all tasks as follows: $bs=20000$ and $top-k=16$. The training parameters are $\alpha=0.25$, $\epsilon=0.0002$, and RGCN embedding-dim=128.
Our SPARQL-based method is based on two parameters, $d \in \{1,2\}$ for edge direction and $h \in \{1, 2,...,n\}$ for the number of hops. We evaluated four variations of {\sysName} where $d$ and $h$ ranged from 1 to 2, as shown in Figure~\ref{fig:DifferentParameters}.}{R3.O1} 

We used {\sysName}$_{dihj}$ to extract $KG'$ and compared its performance against BRW and IBS (our baselines). The parameters in {\sysName}$_{dihj}$ determine the TOSG sampling scope around a target vertex, which affects the size of $KG'$ depending on the KG structure. Among the six NC tasks, {\sysName}$_{d1h1}$ met all characteristics discussed in Section~\ref{sec:chars} while consuming the least time and memory with comparable accuracy. {\sysName} based on the SPARQL-based variations managed to improve the accuracy in PV/MAG-42M and PV/DBLP-15M and maintain comparable accuracy on PC/YAGO-30M while reducing time and memory w.r.t BRW and IBS. \highlightedReply{YAGO-30 showcases diversity with 104 node types and 98 edge types (see Table \ref{tbl_exp_ds_Statistics}). BRW and IBS employ aggressive sampling that improves the identification of critical features (nodes and edges). Our three methods extract TOSGs with varying ratios of $\mathcal{V}_T$ and $\mathcal{V}_s$, as shown in Table III. IBS, for YAGO-30 PC, yields a $KG'$ with an average of 74.6\% target vertices (25.4\% $\mathcal{V}_s$), while BRW produces a $KG'$ with an average of 55.3\% target vertices (54.7\% $\mathcal{V}_s$). Hence, IBS is faster on YAGO-30 PC compared to BRW, whereas BRW demonstrates faster performance than IBS on MAG and DBLP. On the one hand, BRW and IBS achieved better accuracy in the YAGO-30 PC task, but the sampling cost of both methods introduced excessive training time and memory consumption.
On the other hand, {\sysName}$_{d1h1}$ introduces the best balance by achieving comparable or better accuracy in all tasks with negligible preprocessing cost.}{R1.O1} We break down the {\sysName} preprocessing cost in table \ref{tab:preprocessing}. We analyzed the sensitivity of these parameters on the LP tasks, where {\sysName}$_{d2h1}$ fulfils all our evaluation criteria.\shorten

\begin{figure}[t]
\vspace*{-3ex}
  \centering
   \includegraphics [width=0.95\columnwidth ]{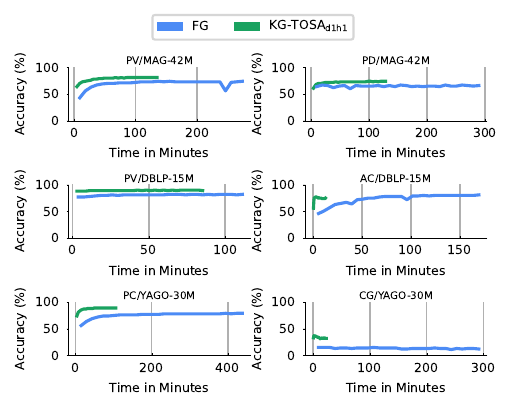}
   \vspace*{-2.5ex}
  \caption{Convergence rate analysis. GraphSAINT while training the six NC tasks using the full graph (FG) and $KG'$ extracted by {\sysName}. {\sysName} enables the GNN method to generalize faster with comparable accuracy.\shorten}
  \label{fig:KGTOSA_Results_Acc_vs_Time}
  \vspace*{-3ex}
\end{figure}

\begin{table*}[t]
\vspace*{-3ex}
\centering
\caption{
A breakdown of the cost involved in training a GNN task: \myNum{i} {\sysName}$_{d1h1}$ to extract $KG'$ from FG (optional), \myNum{ii} transforming the RDF triples into adjacency matrices (mandatory), and \myNum{iii} training the task using GraphSaint. 
% $KG'$ is extracted by {\sysName} with $d=1$ and $h=1$. FG refers to the cost of training on the entire KG, and $KG'$ refers to the cost while using {\sysName}.\shorten 
}
\vspace*{-1ex}
\label{tab:preprocessing}
% \small
\scriptsize
\begin{tabular}{|l|c|c|c|c|c|c|c|c|c|c|c|c|}
\hline
% &\multicolumn{4}{c|}{\textbf{MAG-42M}}& \multicolumn{4}{c|}{\textbf{DBLP-15M}}& \multicolumn{4}{c|}{\textbf{YAGO-30M}} \\\hline 
&\multicolumn{2}{c|}{\textbf{PV/MAG-42M}}&\multicolumn{2}{c|}{\textbf{PD/MAG-42M}}& \multicolumn{2}{c|}{\textbf{PV/DBLP-15M}}&\multicolumn{2}{c|}{\textbf{AC/DBLP-15M}}& \multicolumn{2}{c|}{\textbf{PC/YAGO-30M}}&\multicolumn{2}{c|}{\textbf{CG/YAGO-30M}} \\\hline 
\textbf{Step}&FG&KG'&FG&KG'&FG&KG'&FG&KG'&FG&KG'&FG&KG'\\ \hline
KG Extraction Time (mins)&-&18&-&16&-&19&-&1&-&22&-&3\\ \hline
Transformation Time (mins)&46&22&41&19&30&11&9&1&52&10&60&3\\ \hline
GNN Training Time (mins)&274&135&290&129&112&85&170&13&439&105&292&23\\ \hline
\textbf{Total Time (mins)}&\textbf{320}&{175}&\textbf{331}&{164}&\textbf{142}&{115}&\textbf{179}&{15}&\textbf{491}&{137}&\textbf{352}&{29}\\ \hline\hline
Accuracy(\%)&74&\textbf{81}&67&\textbf{74}&82&\textbf{90}&\textbf{81}&79&79&\textbf{90}&15&\textbf{37}\\ \hline
Model Size (\#Params (M)) &\textbf{5349}&1415&\textbf{5348}&1408&\textbf{3301}&1477&\textbf{3306}&96&\textbf{3656}&1085&\textbf{3933}&1038\\ \hline
Inference Time (sec)&\textbf{89}&52&\textbf{87}&52&\textbf{678}&454&\textbf{1003}&28&\textbf{1265}&368&\textbf{1283}&32\\ \hline
Training Memory (GB)&\textbf{155}&57&\textbf{139}&57&\textbf{47}&36&\textbf{80}&3&\textbf{130}&30&\textbf{90}&3 \\ \hline
\end{tabular}
\vspace*{-2ex}
\end{table*}

\subsubsection{Analyzing our Extraction Methods in terms of Data Sufficiency and Graph Topology}
We analyzed the quality indicators of subgraphs generated using our sampling and SPARQL-based methods against the URW for four different tasks on three KGs, as shown in Table~\ref{tab:tbl_YAGO_compare_stat}. The BRW, IBS, and {\sysName}$_{d1h1}$ subgraphs statistics consistently show better quality indicators, compared to URW. For data sufficiency, the percentage of target nodes is deficient in URW samples, which range from 1\% to 30\%. Our methods increased this ratio in most of the cases. {\sysName}$_{d_1h_1}$ maintained the highest balance between 35\% and 40\% in all cases. Our generic graph pattern helps the three methods to extract relevant node/edge types. However, URW may include irrelevant node/edge types. Hence, the number of $|\mathcal{C}^{\prime}|$ and $|\mathcal{R}^{\prime}|$ is decreased in most of the cases by our three methods.\shorten 

For graph topology, our methods excluded disconnected target nodes, reduced the average distance between non-target and target nodes, and enhanced the average entropy of neighbour node types as calculated by Equation \ref{eq_shannon}. This results in a subgraph with fewer hops and a greater diversity of node types compared to URW.
Thus, training models using the $KG’$ helps reduce GNN oversmoothing~\cite{DropEdge}. Overall, our three methods outperform GraphSAINT, which uses URW by default, with up to 27\% better accuracy. These results demonstrate that our methods allow the extraction of higher-quality training subgraphs (mini-batches). However, IBS and BRW incur an excessive extraction overhead, particularly with a large KG that may overshadow the potential time savings.\shorten

\subsubsection{Task-oriented sampling and Model Convergence Rate}
\label{subsec_convergance_rate}
To analyze the time reduction achieved by {\sysName}, we tracked the accuracy achieved by GraphSAINT as a function of the number of training epochs while using FG and $KG’$. We run with 30 epochs each. 
However, the average time per epoch with $KG’$ is much shorter. 
Thus, we report the time instead of the number of epochs in Figure~\ref{fig:KGTOSA_Results_Acc_vs_Time}. 
Across all tasks, {\sysName} enables GraphSAINT to achieve a faster convergence rate. 
{\sysName} identifies a smaller subset of the KG, i.e., $KG’$, that helps GraphSAINT generate mini-batches of high-quality sub-graphs, as shown in Table~\ref{tab:tbl_YAGO_compare_stat}. Hence, the model attains optimal performance in fewer and faster iterations, which helps reduce training times and computational costs.
We got similar results~\footnote{
Results are shown in the \href{https://github.com/CoDS-GCS/KGTOSA/blob/main/KGTOSA_SupplementalMaterial.pdf}
{\textcolor{blue}{supplementary materials}}  due to the lack of space. \shorten
%\hr,f{https://gitfront.io/r/HGNN/LWYY16iUMSuE/KGTOSA/raw/KGTOSA_SupplementalMaterial.pdf}
} with other GNN methods with an even better convergence rate.\shorten

\subsubsection{The {\sysName} Cost Breakdown in GNN Training} 
GNN methods require a KG to be provided as adjacency matrices.
Thus, traditional GNN pipelines involve data transformation from RDF triples to adjacency matrices. For extracting $KG'$, {\sysName}$_{d1h1}$ used VM$^{TB}$ with $\mathcal{P}= 64$ and a batch size $bs$ of 1M tiples, see Algorithm~\ref{alg:SPARQL_MS}. For the six NC tasks, Table~\ref{tab:preprocessing} summarizes the cost breakdown, accuracy, model size, inference time, and memory usage for traditional GNN pipelines (FG) and the pipeline with {\sysName}$_{d1h1}$ ($KG'$) for our NC tasks. {\sysName} reduces training and data transformation time, which empowers GNN methods to achieve performance gains that outweigh its negligible preprocessing overhead. {\sysName} identifies a subgraph, $KG'$, of smaller size and structure. That helps models trained using {\sysName} exhibit a remarkable reduction in size, up to 34 times smaller, while demonstrating significantly faster inference times, up to 40 times faster, as shown in Table~\ref{tab:preprocessing}.\shorten 
\section{Related Work}
\label{sec:relatedwork} 

\highlightedReply{We developed {\sysName} as part of our vision towards~\cite{KGNet,SPARQL_ML} an on-demand graph ML (GML) as a service on top of KG engines to support GML-enabled SPARQL queries. The {\sysName} task-oriented sampling aims to prune the search space by identifying the minimal subgraph $sg$ in a KG, where $sg$ is related to the task at hand.}{R3.O2} Unlike our approach, IBMB~\cite{ibmb} introduced 
Personalized Pagerank (PPR) to calculate the influenced scores in homogeneous graphs to sample a representative graph that could be used to train different tasks. IBMB does not support KGs and task-oriented HGNNs training.\shorten

There is a growing effort to develop systems for training GNNs on large graphs at scale. These systems aim to \myNum{i} distribute the training across multiple GPUs or CPUs in a single machine (scale-up)~\cite{NeuGraph, G3}, or \myNum{ii} distribute the training across a cluster of machines with multiple GPUs or CPUs per machine (scale-out)~\cite{DistDGL, DGCL, Sancus, AliGraph,TF-GNN}.  {\sysName} can be used with scale-out and scale-up systems to optimize the use of existing resources.  
All the used GNN methods in our evaluation are based on scale-up frameworks, such as the PYG~\cite{pytorch-geometric}. NeuGraph~\cite{NeuGraph} leverages multi-GPU in single-machine to improve training performance. 
% G3~\cite{G3} uses parallel graph optimizations to improve graph operations in GPU systems. 
Marius~\cite{Marius} optimizes graph embedding learning by using partition caching and buffer-aware data ordering.
There are several scale-out GNN frameworks, as follows. DistDGL~\cite{DistDGL} uses partitioning with load balancing to achieve high scalability. 
% DGCL~\cite{DGCL} reduces communication overhead by finding optimal communication routes based on graph structure. 
SANCUS~\cite{Sancus} reduces communication overhead using a powerful parallel algorithm and abstracting GNN processing as sequential matrix multiplication.
% This allows it to cache intermediate historical embeddings and reuse them during training.
AliGraph~\cite{AliGraph} also uses static cache but only supports CPU servers. 
% AGL\cite{AGL}  uses MapReduce and optimizes both training and inference. 
%Finally, TF-GNN\cite{TF-GNN} is designed to allow training on graph structures that do not fit into memory by supporting many partitioning strategies via METIS. 
DGL-Dist\cite{DistDGL} supports RGCN and different sampling techniques.\shorten

% Existing GNN benchmarks in \cite{OGB, OGB-LSC, HGNN_Benchmark} contain either large homogeneous graph datasets or heterogeneous datasets with a few numbers of node/edge types such as MAG240M\cite{OGB-LSC} with three node and edge types and 1.7B Billion triples. Large heterogeneous datasets especially KGs are used for link prediction tasks using KGE methods \cite{KGE_Survey}. Real KGs contain hundreds to thousands of node and edge types such as Wikidata \cite{KG_Wikidata} containing 1,301 and 10,012 node and edge types and Yago~\cite{KG_Yago4} containing 8,902 and 156 node and edge types, respectively. There is a need to challenge the existing HGNN methods in real KG settings.  

\vspace*{0ex}
\section{Conclusion}
\label{sec:conclusion} 

Handcrafted task-oriented subgraphs (TOSG) help GNN methods reduce training and memory usage by trading accuracy for performance. Extracting a TOSG from an arbitrary KG is time-consuming and challenging for AI practitioners. In this paper, we conducted a comprehensive study to analyze these challenges. Based on our analysis, we discovered a generic graph pattern that captures local and global KG structures relevant to training a specific task. 
This paper proposes two task-oriented sampling techniques based on biased random walks and Personalized PageRank scores for extracting subgraphs matching our generic graph pattern. To address the computational overhead of the task-oriented sampling techniques, we propose a SPARQL-based method leveraging RDF engines for TOSG extraction. This method also avoids the full KG migration while achieving comparable or better modelling performance.
This paper benchmarks our approach on large KGs, such as MAG, DBLP, YAGO, and Wikidata,  with diverse node classification and link prediction tasks. Our comprehensive evaluation includes six state-of-the-art GNN methods. Our results demonstrate that our approach significantly reduces training, inference time and memory footprint while improving performance metrics (e.g., accuracy, Hits@10). Our approach emerges as a cost-effective HGNN training approach, particularly beneficial for large KGs. This is a step forward to developing sustainable approaches and optimizations for data science.

\nsstitle{Acknowledgement.}
We express our gratitude to Dr. Nesreen K. Ahmed for her valuable feedback on our approach and conducted experiments. Additionally, we extend our thanks to Dr. Mikhail Galkin for his thorough review of an earlier version of this paper.  
\newpage

\balance
\bibliographystyle{IEEEtran}
\bibliography{refrences}
\end{document}